\documentclass{article} %
\usepackage{src/preprint,times}

\usepackage{amsmath,amsfonts,bm}

\def\eqref#1{equation~\ref{#1}}

\def\1{\bm{1}}

\DeclareMathAlphabet{\mathsfit}{\encodingdefault}{\sfdefault}{m}{sl}
\SetMathAlphabet{\mathsfit}{bold}{\encodingdefault}{\sfdefault}{bx}{n}

\usepackage{hyperref}
\hypersetup{colorlinks=true,linkcolor=[rgb]{0.6,0.1,0.1},citecolor=[rgb]{0.1,0.3,0.6},urlcolor=[rgb]{0.1,0.3,0.6}}
\usepackage{url}
\usepackage{tikz}
\usepackage{booktabs}
\usepackage{longtable}

\usepackage{caption}
\usetikzlibrary{positioning,arrows.meta,calc,decorations.pathreplacing}
\title{The Behavioral Rate of Fine-Tuning:
Compression as a Measure of Useful Information}

\author{Joey David$^{1}$, Paul Caillon$^{1}$, Alexandre Allauzen$^{1,2}$ \\
$^1$ Miles Team, LAMSADE, Université Paris Dauphine-PSL, Paris, France \\
$^2$ ESPCI PSL, Paris, France \\
{\small\texttt{joey.david@dauphine.eu},\,\texttt{\{paul.caillon,alexandre.allauzen\}@dauphine.psl.eu}}
}

\iclrfinalcopy %
\begin{document}

\maketitle

\begin{abstract}
Low-Rank Adaptation (LoRA) adapts a frozen language model through a compact learned update, making it a natural setting for asking how much information fine-tuning needs to write.
Yet adapter rank, parameter count, or numerical reconstruction error do not directly tell us how many bits are required to preserve the acquired behavior.
We therefore define an operational behavioral rate $R^\star$ as the smallest serialized size of a compressed adapter that, once reloaded into the frozen base model, retains a fixed fraction of its held-out fine-tuning gain.
Across LoRA fine-tuning experiments, this rate increases with distinct training content at fixed optimization budgets, but is not determined by intuitive metrics like source description length, fine-tuning gain, or weight-space reconstruction error.
Most importantly, the same datasets exhibit substantially different, and even reordered, rates across frozen base models, showing that fine-tuning compressibility is a property of the dataset--receiver pair rather than of the dataset alone.
However, we show that correction spectra measured on the data are able to rank these rates within a receiver.
Finally, we study a striking noise-reducing effect of compression: adapters trained on deliberately corrupted datasets fall far below the frozen model, yet after compression score above it (23.7\% to 82.9\% on GSM8K, against 77.6\% for frozen Qwen2.5-7B), consistently across tested families and sizes.
The damage is unevenly concentrated in the adapter's leading singular directions, which compression appears to filter.
Away from its training task, the compressed corrupted adapter also generalizes better than a clean fine-tune: on harder math it retains 97\% of base accuracy, versus 44\% for the clean adapter.
Together, these results suggest that the information cost of fine-tuning depends on the receiver, target behavior, and how the required corrections are represented in the learned update.
\end{abstract}

\section{Introduction}

Fine-tuning changes the behavior of a pretrained model by writing an update
relative to what that model already knows. How many bits are actually required
to preserve this acquired behavior? Parameter count, adapter rank, and numerical
precision measure storage, but need not measure the information that matters
functionally. Low-Rank Adaptation (LoRA) \citep{hu2022lora} provides a clean
setting for studying this question: the base model remains frozen while the
learned update is isolated in a separate adapter that can be compressed after
training.
\newline
We take a receiver-relative rate--distortion view of this process. As
illustrated in Figure~\ref{fig:information_view}, fine-tuning acts as an
encoder, the adapter as a finite-rate message, and the frozen model as side
information available when that message is decoded. We measure rate by the
exact serialized size of a compressed adapter and distortion by the behavior
of the resulting composed model. For a fixed retention target, the
\emph{behavioral rate} $R^\star$ is the smallest measured rate that preserves
the required fraction of the held-out fine-tuning gain.
\newline
This operational definition exposes a gap between numerical and behavioral
compression. Lower weight reconstruction error need not preserve more learned
behavior, and comparable bit budgets can have different effects depending on
where compression acts. Behavioral rate also increases with distinct training
content at fixed optimization budget. Most importantly, the same datasets can
require substantially different rates across frozen models, with their
ordering changing across receivers. A correction-gradient spectrum computed
from the data and receiver ranks $R^\star$ well within a fixed receiver, but
does not transfer reliably across new receivers.
\newline
Compression can also reveal structure within a learned update. Under
deliberately corrupted reasoning supervision, fine-tuning can strongly damage
the frozen model, yet aggressive compression recovers much of the lost
performance and can exceed the frozen baseline. On Qwen2.5-7B, for example,
GSM8K accuracy falls from 77.6\% to 23.7\% after corrupted fine-tuning and
returns to 82.9\% after compression. Singular-direction analyses show that
harmful and useful effects are distributed unevenly across the update, while
matched-norm controls show that the recovery is not exactly equivalent to
scalar rescaling.
\newline
Finally, once an update has been trained, its behavioral rate closely tracks
how far it can be shrunk before losing its gain. A simple codec-specific rule,
$R^\star(0.90)\approx\alpha_{90}/\gamma$, achieves an RMSE of 0.068 bits per
value on 42 checkpoints from five receivers not used to fix the rule, although
the evaluation remains retrospective. Together, these results suggest that
the information cost of fine-tuning is not an intrinsic property of the data
or of parameter displacement alone: it depends on the frozen receiver, the
behavior to preserve, and how the learned corrections are represented within
the update.
\definecolor{srcFill}{HTML}{EEF7EE}
\definecolor{srcDraw}{HTML}{8BCF8B}
\definecolor{srcIcon}{HTML}{3F8F3F}
\definecolor{encFill}{HTML}{F5EAF3}
\definecolor{encDraw}{HTML}{C28FB9}
\definecolor{encIcon}{HTML}{9A4D8E}
\definecolor{msgFill}{HTML}{E6EEF7}
\definecolor{msgDraw}{HTML}{3775BA}
\definecolor{msgIcon}{HTML}{0F4D92}
\definecolor{baseFill}{HTML}{F2F2F2}
\definecolor{decDraw}{HTML}{4D4D4D}
\definecolor{baseIcon}{HTML}{767676}
\definecolor{adpFill}{HTML}{E6EEF7}
\definecolor{adpIcon}{HTML}{0F4D92}
\definecolor{outFill}{HTML}{E5F2F3}
\definecolor{outDraw}{HTML}{42949E}
\definecolor{outIcon}{HTML}{2C6F77}
\definecolor{distDraw}{HTML}{B64342}

\begin{figure}[t]
\centering
\resizebox{\linewidth}{!}{%
\begin{tikzpicture}[
  font=\normalsize,
  >={Latex[length=2.7mm,width=2.0mm]},
  flow/.style={->,line width=1.15pt,draw=decDraw},
  panel/.style={
    rounded corners=2.5pt,
    line width=1.0pt,
    minimum width=3.15cm,
    minimum height=2.85cm,
    inner sep=0pt
  },
  role/.style={font=\bfseries\large},
  main/.style={font=\large,align=center,text=black!92},
  sub/.style={font=\itshape\normalsize}
]

\node[panel,draw=srcDraw,fill=srcFill] (data) at (0,0) {};
\node[panel,draw=encDraw,fill=encFill] (training) at (4.05,0) {};
\node[panel,draw=msgDraw,fill=msgFill] (adapter) at (8.10,0) {};
\node[minimum width=4.80cm,minimum height=2.85cm,inner sep=0pt] (decoder) at (13.10,0) {};
\node[panel,draw=outDraw,fill=outFill] (behavior) at (18.15,0) {};

\begin{scope}[
    shift={($(data.center)+(0,0.64)$)},
    draw=srcIcon,
    line width=1.15pt,
    scale=1.18
]

\draw[
    fill=srcFill,
    rounded corners=1.2pt
]
(-0.02,-0.28) rectangle (0.34,0.30);

\draw[
    fill=srcFill,
    rounded corners=1.2pt
]
(-0.12,-0.20) rectangle (0.24,0.38);

\draw[
    fill=srcFill,
    rounded corners=1.2pt
]
(-0.22,-0.12) rectangle (0.14,0.46);

\draw[line width=.9pt] (-0.14,0.26) -- (0.06,0.26);
\draw[line width=.9pt] (-0.14,0.13) -- (0.06,0.13);
\draw[line width=.9pt] (-0.14,0.00) -- (0.02,0.00);

\end{scope}
\node[main] at ($(data.center)+(0,-0.70)$) {Fine-tuning\\data};

\begin{scope}[shift={($(training.center)+(0,0.66)$)},scale=1.18]
  \foreach \a in {0,45,...,315}{\begin{scope}[rotate=\a]\fill[encIcon] (-0.08,0.29) rectangle (0.08,0.43);\end{scope}}
  \fill[encIcon] (0,0) circle (0.34);
  \fill[encFill] (0,0) circle (0.13);
\end{scope}
\node[main] at ($(training.center)+(0,-0.62)$) {Fine-tuning};

\begin{scope}[shift={($(adapter.center)+(0,0.66)$)},draw=msgIcon,line width=1.25pt,scale=1.18]
  \draw[fill=msgFill,rounded corners=2.6pt] (-0.34,-0.31) rectangle (0.40,0.31);
  \foreach \yy in {-0.18,0,0.18}{\draw (-0.50,\yy)--(-0.34,\yy);}
  \draw (0.02,-0.17)--(0.02,0.17);
  \draw (0.14,-0.17)--(0.14,0.17);
  \draw (0.26,-0.17)--(0.26,0.17);
\end{scope}
\node[main] at ($(adapter.center)+(0,-0.70)$) {Learned\\adapter};

\begin{scope}
  \clip[rounded corners=2.5pt] (decoder.south west) rectangle (decoder.north east);
  \fill[baseFill] (decoder.south west) rectangle (decoder.center |- decoder.north);
  \fill[adpFill] (decoder.center |- decoder.south) rectangle (decoder.north east);
\end{scope}
\draw[draw=decDraw,line width=1.0pt,rounded corners=2.5pt]
  (decoder.south west) rectangle (decoder.north east);

\draw[dashed,dash pattern=on 3pt off 2.4pt,draw=black!34,line width=.72pt]
  ($(decoder.center)+(0,1.425)$)--($(decoder.center)+(0,0.38)$);
\draw[draw=black!34,line width=.72pt]
  ($(decoder.center)+(0,0.38)$) arc[start angle=90,end angle=-90,radius=.20cm];
\draw[dashed,dash pattern=on 3pt off 2.4pt,draw=black!34,line width=.72pt]
  ($(decoder.center)+(0,-0.38)$)--($(decoder.center)+(0,-1.425)$);

\begin{scope}[
    shift={($(decoder.west)!0.5!(decoder.center)+(0,0.49)$)},
    x=0.34cm,
    y=0.27cm
]

\foreach \yi in {-1,0,1}{
    \foreach \yh in {-1.35,-0.45,0.45,1.35}{
        \draw[
            draw=baseIcon!55,
            line width=0.45pt
        ] (-1,\yi) -- (0,\yh);
    }
}

\foreach \yh in {-1.35,-0.45,0.45,1.35}{
    \foreach \yo in {-1,0,1}{
        \draw[
            draw=baseIcon!55,
            line width=0.45pt
        ] (0,\yh) -- (1,\yo);
    }
}

\foreach \y in {-1,0,1}{
    \filldraw[
        fill=baseFill,
        draw=baseIcon,
        line width=0.75pt
    ] (-1,\y) circle (0.105cm);

    \filldraw[
        fill=baseFill,
        draw=baseIcon,
        line width=0.75pt
    ] (1,\y) circle (0.105cm);
}

\foreach \y in {-1.35,-0.45,0.45,1.35}{
    \filldraw[
        fill=baseFill,
        draw=baseIcon,
        line width=0.75pt
    ] (0,\y) circle (0.105cm);
}

\end{scope}
\node[font=\large,text=black!92] at ($(decoder.west)!0.5!(decoder.center)+(0,-0.48)$) {Base model};
\node[sub,text=baseIcon] at ($(decoder.west)!0.5!(decoder.center)+(0,-0.91)$) {frozen};

\begin{scope}[shift={($(decoder.center)!0.5!(decoder.east)+(0,0.70)$)},draw=adpIcon,line width=1.25pt,scale=1.18]
  \draw[fill=adpFill,rounded corners=2.6pt] (-0.34,-0.31) rectangle (0.40,0.31);
  \foreach \yy in {-0.18,0,0.18}{\draw (-0.50,\yy)--(-0.34,\yy);}
  \draw (0.02,-0.17)--(0.02,0.17);
  \draw (0.14,-0.17)--(0.14,0.17);
  \draw (0.26,-0.17)--(0.26,0.17);
\end{scope}
\node[font=\large,text=black!92] at ($(decoder.center)!0.5!(decoder.east)+(0,-0.48)$) {Adapter};
\node[sub,text=adpIcon] at ($(decoder.center)!0.5!(decoder.east)+(0,-0.91)$) {trained};

\begin{scope}[shift={($(behavior.center)+(0,0.66)$)},draw=outIcon,line width=1.25pt,scale=1.18]
  \draw (0,0) circle (.34);
  \draw (0,0) circle (.21);
  \draw (0,0) circle (.07);
  \draw[->] (.39,.39)--(.0,.0);
\end{scope}
\node[main] at ($(behavior.center)+(0,-0.70)$) {Adapted\\behavior};

\node[role,text=srcIcon] at ($(data.north)+(0,1.00)$) {SOURCE};
\node[role,text=encIcon] at ($(training.north)+(0,1.00)$) {ENCODER};
\node[role,text=msgIcon] at ($(adapter.north)+(0,1.00)$) {MESSAGE};
\node[role,text=decDraw] at ($(decoder.north)+(0,1.00)$) {DECODER};
\node[role,text=outIcon] at ($(behavior.north)+(0,1.00)$) {RECONSTRUCTION};

\foreach \node/\col in {data/srcIcon,training/encIcon,adapter/msgIcon,decoder/decDraw,behavior/outIcon}{
  \draw[\col,line width=.75pt,rounded corners=2pt]
    ($ (\node.north west)+(0,0.38) $)
    -- ($ (\node.north west)+(0,0.50) $)
    -- ($ (\node.north east)+(0,0.50) $)
    -- ($ (\node.north east)+(0,0.38) $);
}

\draw[flow] (data.east)--(training.west);
\draw[flow] (training.east)--(adapter.west);
\draw[flow] (adapter.east)--(decoder.west);
\draw[flow] (decoder.east)--(behavior.west);

\draw[decorate,decoration={brace,mirror,amplitude=4.4pt},draw=msgIcon,line width=.8pt]
  ($(adapter.south west)+(0,-.12)$)--($(adapter.south east)+(0,-.12)$);
\node[font=\large,text=msgIcon] (rate) at ($(adapter.south)+(0,-.62)$) {rate $R$};

\coordinate (dbot) at ($(adapter.south)+(0,-1.46)$);
\draw[->,dash pattern=on 4pt off 3pt,line width=.9pt,draw=distDraw]
  (data.south)
  -- (data.south |- dbot)
  -- node[pos=.53,below=5pt,font=\large,text=distDraw,fill=white,inner xsep=2pt]
     {behavioral distortion $D$}
  (behavior.south |- dbot)
  -- (behavior.south);

\end{tikzpicture}%
}

\caption{Information-theoretic view of parameter-efficient fine-tuning. $R$ is the number of bits used by the adapter, $D$ measures failure to preserve the behavior induced by fine-tuning.}
\label{fig:information_view}
\end{figure}
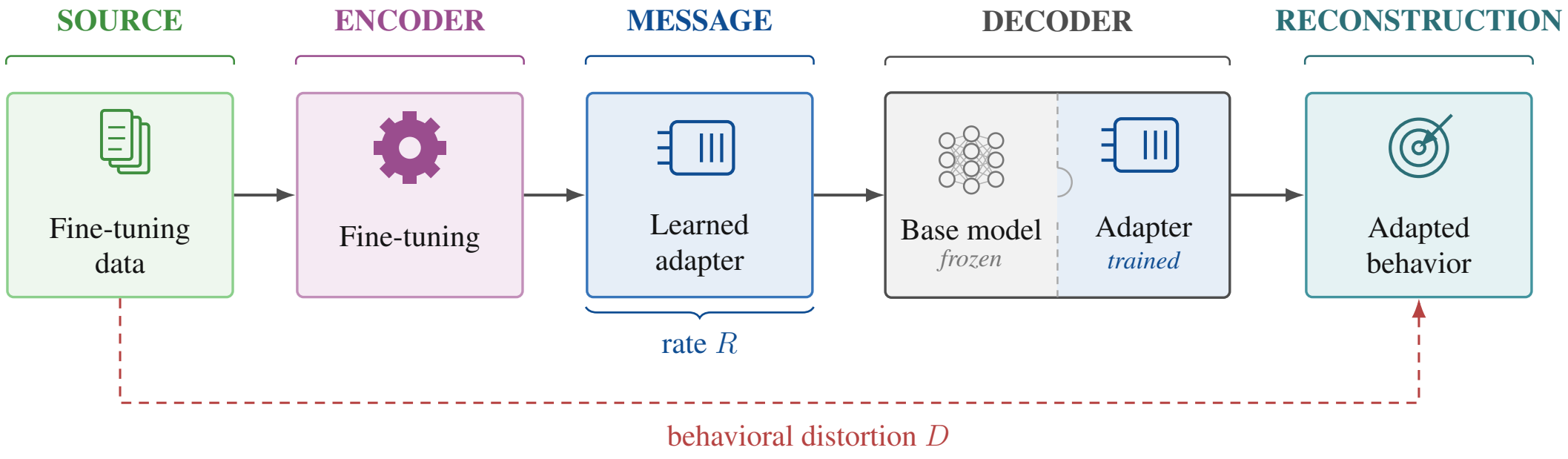

\section{Related Work}

\textbf{Parameter-efficient fine-tuning and adapter structure.}
LoRA \citep{hu2022lora} and QLoRA \citep{dettmers2023qlora} provide compact
parameterizations for adapting frozen language models, while earlier work on
intrinsic dimensionality showed that effective fine-tuning can lie in a
low-dimensional subspace \citep{aghajanyan2021intrinsicdim}. More recent work
has emphasized that the components of a learned adapter need not be
functionally equivalent. \citet{zhu2024loraasymmetry} study asymmetric roles
within low-rank adapters, \citet{shuttleworth2025intruder} identify
high-ranking singular directions introduced by LoRA whose attenuation can
reduce forgetting, and \citet{tian2026spectralsurgery} identify neutral or
harmful singular directions that can be reweighted using calibration
gradients. Our work is complementary: we use post-training compression to
probe which components of an update are required to preserve its acquired
behavior.

\textbf{Quantization and adapter compression.}
Low-bit post-training quantization has been extensively studied for language
models, including ZeroQuant \citep{yao2022zeroquant}, GPTQ
\citep{frantar2023gptq}, SmoothQuant \citep{xiao2023smoothquant}, and AWQ
\citep{lin2024awq}; more recent work characterizes behavior in extremely
low-bit regimes \citep{liu2025paretoq}. Quantization has also been extended
to trained adapters: \citet{mirzaei2025loraquant} compress LoRA updates to
ultra-low precision while preserving task performance, while
\citet{tan2026adapterbits} study adapter capacity and memorization directly
in bits. In contrast, we use compression primarily as a measurement tool:
our rate is the serialized size required to preserve the behavior induced by
fine-tuning.

\textbf{Compression, description length, and rate--distortion.}
Model compressibility has been connected to generalization and description
length \citep{arora2018genbounds,lotfi2022pacbayescomp,
urdshals2025compressibility}, and rate--distortion ideas have been applied to
language-model compression \citep{young2025radio}. Our setting differs in two
respects: the frozen model acts as shared side information, and distortion is
measured behaviorally after the compressed adapter is decoded back into that
receiver. The resulting rate is therefore relative to both the learned update
and the frozen model with which it is composed.

\section{Functional Structure of Adapter Compression}
\label{sec:diagnostics}

We begin by examining how post-training compression affects fine-tuned
behavior. Controlled compression lets us separate numerical reconstruction
from functional preservation while keeping both the learned checkpoint and
the frozen model fixed. These diagnostics show that ordinary parameter-space
fidelity is not sufficient to predict which compressed adapters preserve the
behavior acquired during fine-tuning.

\subsection{Experimental Setting: Adapter Compression}

We study post-training compression of LoRA adapters across frozen language
models from 0.5B to 32B parameters, including
Qwen2.5-\{0.5B, 1.5B, 3B, 7B, 14B, 32B\}, Qwen3-8B-Base,
Mistral-7B-v0.1, Llama-3.1-8B, and Gemma-2-9B. 
Each adapter is trained once and then compressed over a range of operating
points, without retraining. At one bit per value and above, we quantize the
LoRA factors row-wise; below one bit, we additionally remove paired rank
directions. Each compressed representation, including its decoding metadata,
is serialized to disk, reloaded, and evaluated in composition with the frozen
model. We report realized file size normalized by the number of scalar values
in the uncompressed adapter. Full training configurations, codec details,
datasets, and evaluation procedures are given in
Appendix~\ref{app:xp_details}.

\subsection{Measuring Compression Through Behavior}

\begin{figure}[t]
\centering
\includegraphics[width=0.61\linewidth]{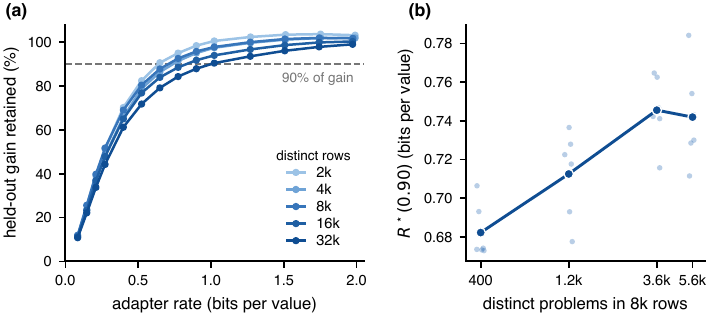}
\hfill
\includegraphics[width=0.36\linewidth]{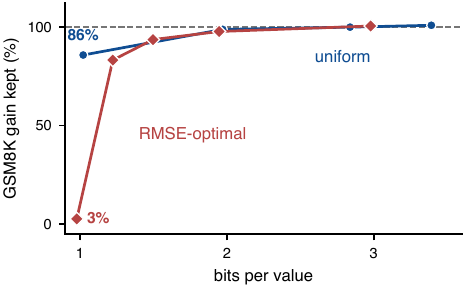}
\caption{
\textbf{Behavioral rate is not a weight-space quantity.}
\textbf{Left:} retained fine-tuning gain versus exact serialized adapter rate
across 2k--32k MetaMathQA rows at fixed compute.
\textbf{Center:} $R^\star(0.90)$ at 8k rows as source-problem diversity
increases.
\textbf{Right:} near one bit per value, an RMSE-optimized codec achieves lower
relative weight error than uniform quantization (0.54 vs.\ 0.56) but retains
only 3\% of the GSM8K gain, versus 86\%.
Curves on the left show seed means; small points are individual seeds.
Mistral-7B, NF4, rank-16 LoRA, three seeds.
}
\label{fig:compression-diagnostics}
\end{figure}

The object we ultimately want to preserve is not the adapter itself, but the
behavior added by fine-tuning. Because every operating point is obtained from
the same trained checkpoint, differences after compression isolate the effect
of representation from variation in optimization.
\newline
Let $L_0$ denote the held-out loss of the frozen model,
$L_{\mathrm{raw}}$ that of the uncompressed adapter, and $L_r$ the loss after
compression to realized rate $r$. We define retained gain as
\begin{equation}
    \mathrm{Retain}(r)
    =
    \frac{L_0-L_r}{L_0-L_{\mathrm{raw}}},
    \label{eq:retain}
\end{equation}
so that $0$ corresponds to the frozen model and $1$ to the uncompressed
adapter; values above one are possible when compression improves held-out
loss. For comparisons across conditions, we define
\begin{equation}
    R^\star(\tau)
    =
    \min\{r:\mathrm{Retain}(r)\geq\tau\},
    \label{eq:rstar}
\end{equation}
and use $\tau=0.90$ unless stated otherwise. $R^\star(0.90)$ is therefore an
operational, codec-dependent behavioral rate: the smallest measured reloadable
adapter preserving at least 90\% of the original fine-tuning gain.
\newline
Figure~\ref{fig:compression-diagnostics} (left and center) illustrates the
resulting rate-retention frontier. At fixed optimization budget, more
distinct training content shifts the frontier toward higher rates; at fixed
dataset size, $R^\star(0.90)$ also tends to increase with source-problem
diversity. We return to what determines these shifts in
Section~\ref{sec:behavior-prediction}.
\newline
A natural alternative would be to measure preservation through numerical
reconstruction error. We test this by comparing uniform quantization with an
adaptive allocator that minimizes LoRA weight reconstruction error under a
rate constraint. Figure~\ref{fig:compression-diagnostics} (right) gives a
direct counterexample: near one bit per value, the adaptive code achieves
lower relative weight RMSE than uniform quantization (0.54 versus 0.56), yet
retains only 3\% of the held-out gain, compared with 86\% for the uniform
code. The allocator partly obtains its lower error by removing rows that are
inexpensive under the Euclidean objective but behaviorally important; at
higher rates, once rows are no longer removed, the two codecs converge.
\newline
Thus, lower parameter-space distortion need not imply lower behavioral
distortion. Numerical reconstruction alone does not identify which
perturbations are harmless once the adapter is composed with the frozen model.
Additional codec details and controls are given in
Appendix~\ref{app:xp_details}.

\section{A Receiver-Relative Rate-Distortion View of Fine-Tuning}
\label{sec:information_theory}

This section develops a receiver-relative rate--distortion view of fine-tuning.
Our goal is not to derive an exact information-theoretic law for LoRA, but to
formalize the empirical observations of Section~\ref{sec:diagnostics} and
motivate the quantities studied later in the paper. We first treat the frozen
model as shared side information, then characterize behavioral preservation
through a functional geometry over adapter perturbations, and finally show how
compression can improve behavior when the learned update is misaligned with
the target utility. Detailed derivations and assumptions are given in
Appendix~\ref{app:rate-distortion}.

\subsection{The Frozen Model as Side Information}

A fine-tuning adapter is not used independently from the model on which it was trained.
At inference time, its parameters act through their composition with the frozen base model.
As illustrated in Figure~\ref{fig:information_view}, the adapter can therefore be viewed as a finite-rate message whose effect is interpreted by a particular receiver.
Let $M$ denote a frozen model, $A$ a trained adapter, and $c$ a finite code of length $\ell(c)$ from which a shared decoder reconstructs
$\hat A = \mathrm{Dec}(c).$
The reconstructed model is $M\oplus\hat A$, where $\oplus$ denotes composition of the frozen model with its adapter.
Since $M$ is available at both encoding and decoding time, it acts as shared side information: the code only needs to specify the additional modification required by this receiver.
\newline
Compression should therefore be evaluated through the behavior of the composed model.
For a utility $U$ for which larger values are better, we define
\(
    D_{M,U}(A,\hat A)
    =
    \left[
        U(M\oplus A)-U(M\oplus\hat A)
    \right]_+ ,
\)
with the corresponding difference reversed for a loss.
Given a tolerance $\varepsilon$, we define the \emph{receiver-relative behavioral description length}
\begin{equation}
    L^\star_{M,U}(A,\varepsilon)
    =
    \min_c \ell(c)
    \quad
    \mathrm{s.t.}
    \quad
    D_{M,U}\!\left(A,\mathrm{Dec}(c)\right)
    \leq \varepsilon .
    \label{eq:behavior-preserving-length}
\end{equation}

Equation~\ref{eq:behavior-preserving-length} makes explicit two dependencies:
changing $M$ changes how an update is interpreted, while changing $U$ changes
which differences between decoded updates matter. This is an operational
one-shot coding problem for a deterministic checkpoint, not a Shannon
rate-distortion function; the empirical rates studied here are therefore
codec-dependent achievable rates rather than information-theoretic optima.

\subsection{Behavioral Distortion Induces a Functional Geometry}

The definition above does not yet specify which parameter perturbations are
costly. Let a compressed adapter satisfy $\hat A=A+e$, where $e$ is the
compression error. Under a local expansion of the utility $U$, and when the
first-order term is negligible for the perturbations considered, the leading
decrease in utility can be written as
\begin{equation}
    D_{M,U}(A,A+e)
    \approx
    \frac{1}{2}e^\top H_U e ,
    \label{eq:functional-quadratic-distortion}
\end{equation}
where $H_U\succeq 0$ denotes minus the local utility Hessian, or equivalently
the local curvature of the corresponding loss near a local optimum.
\newline
This differs from ordinary weight reconstruction error,
$
    D_{\mathrm{weight}}(A,A+e)=\|e\|_2^2,
$
which assigns the same cost to every parameter direction.
Equation~\ref{eq:functional-quadratic-distortion}, in contrast, weights an error by its orientation relative to the behavioral sensitivity of the composed model.
Euclidean and functional distortion induce the same ordering over perturbations only in the isotropic case, when $H_U$ is proportional to the identity on the subspace explored by compression.
\newline
This gives a local interpretation of the RMSE counterexample in
Section~\ref{sec:diagnostics}: behavioral cost depends not only on the
magnitude of a compression error, but on its orientation relative to the
receiver-induced geometry. The matched-budget layer-location controls in
Appendix~\ref{app:layer-location} provide complementary evidence that this
geometry is strongly anisotropic across the network.

\subsection{Why Compression Can Improve on the Base Model}
\label{sec:why-compression-helps}

Equation~\ref{eq:functional-quadratic-distortion} assumes that the first-order term vanishes, i.e.\ that the trained adapter sits at an optimum of the behavior $U$ we care about.
Training only guarantees an optimum of the training loss.
When the data ask for corrections that $U$ does not reward, as with the corrupted supervision of Section~\ref{sec:recovery}, the two optima differ.
Consider moving along the trained update, $M\oplus\alpha A$ for $\alpha\in[0,1]$, so that $\alpha=0$ is the frozen model and $\alpha=1$ the adapter.
To second order,
\begin{equation}
    U(M\oplus\alpha A)
    \approx
    U(M)+\alpha\, g_U^\top A-\frac{\alpha^2}{2}\,A^\top H_U A ,
    \label{eq:overshoot}
\end{equation}
with $g_U$ the gradient of $U$ at the frozen model.
If the update first helps ($g_U^\top A>0$) but is too large, $U$ peaks at $\alpha^\star=g_U^\top A/A^\top H_U A<1$, and every scale between $0$ and $2\alpha^\star$ beats the frozen model while the full update may fall far below it.
A code that effectively moves the update toward such an intermediate scale
can then score above both.
The same argument applies direction by direction: if the harmful excess lives in a few strong directions of $A$, cutting or shrinking those directions recovers the gain the rest of the update carries.
\newline

Compression-based generalization bounds motivate asking whether shorter
adapter descriptions also induce smaller off-task changes
\citep{arora2018genbounds,lotfi2022pacbayescomp}. We evaluate this possibility
in Appendix~\ref{app:recovery-breadth}, where the compressed corrupted adapter
remains substantially closer to the frozen model on off-task probes than a
clean fine-tune. This pattern is consistent with compression suppressing task-specific
perturbations while preserving the useful part of the learned update, although
it does not by itself establish a general relation between shorter adapter
descriptions and better out-of-distribution behavior.
\newline
The same framework also suggests quantities that might anticipate compression
difficulty before training. The per-example correction gradients
$g_i=\nabla_\theta\ell_i(\theta_0)$ describe what the data ask of a particular
receiver. Their second-moment geometry yields a spectrum whose log-determinant
is analogous to the rate term appearing in classical quadratic
rate-distortion models \citep{cover2006elements}. We derive this proxy in
Appendix~\ref{app:correction-statistics} and test its empirical scope in
Section~\ref{sec:behavior-prediction}.

\section{Recovering Task-Relevant Structure through Compression}
\label{sec:recovery}

So far, compression has been treated as a source of distortion. We now ask
whether its non-uniform action can instead be useful when the learned update
itself contains harmful structure.
We construct deliberately corrupted fine-tunes by modifying the supervision
used during training. Our main intervention pairs each MetaMathQA question
with the worked rationale of another example while retaining its original
final answer. The resulting adapter is therefore trained on the correct final
answer together with reasoning that is systematically mismatched to the input.
This sharply degrades task performance and provides a controlled setting in
which useful task signal and harmful supervision can coexist within the same
learned update.

\subsection{Compression Repairs Corrupted Fine-Tunes}

\begin{figure}[t]
\centering
\includegraphics[width=0.55\linewidth]{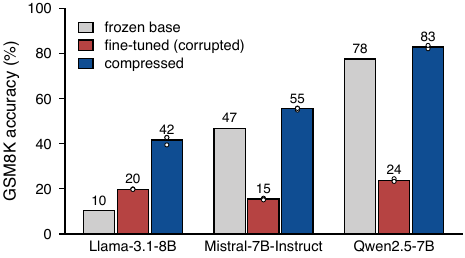}
\caption{
\textbf{Compression can repair a corrupted fine-tune.}
GSM8K accuracy of the frozen model, an adapter trained on permuted rationales,
and the same adapter after lossy compression, across three base models.
}
\label{fig:families}
\end{figure}

Figure~\ref{fig:families} shows that aggressive compression can reverse much
of this damage across several base models. On Qwen2.5-7B, the corrupted
adapter reduces GSM8K accuracy from 77.6\% to 23.7\%; after lossy compression,
the same learned update reaches 82.9\%. The effect also appears on
Mistral-7B-Instruct and Llama-3.1-8B, despite their different frozen-model
baselines.

\begin{figure}[t]
\centering
\includegraphics[width=0.6\linewidth]{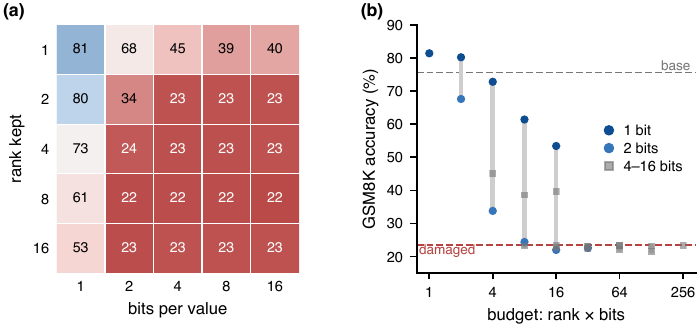}
\caption{
\textbf{Rank and precision are not interchangeable under corrupted
fine-tuning.}
(a) GSM8K accuracy after retaining rank $r$ and coding each value at $b$ bits.
(b) The same configurations grouped by nominal budget $r\times b$.
}
\label{fig:rank-bits}
\end{figure}

Recovery depends strongly on how the compression budget is spent.
Figure~\ref{fig:rank-bits} varies both the number of retained rank directions
and their precision for the same corrupted rank-16 adapter. At nominal budgets
$r\times b\leq16$, retaining more directions at one bit consistently
outperforms retaining fewer directions at higher precision. Most strikingly,
the rank-1 truncation reaches 81.4\% accuracy at one bit, but only 39.6\% at
16 bits. Thus, recovery cannot be explained by parameter count or nominal
budget alone.

A substantial part of the effect is associated with shrinkage: rescaling the
full-precision rank-1 update recovers most of the accuracy obtained by one-bit
coding. Matched-norm controls nevertheless leave a smaller residual advantage
for coarse coding, indicating that norm reduction does not explain the entire
effect. Recovery is also not specific to selecting a single leading singular
direction: full-rank one-bit coding and random low-rank projections can recover
substantial performance as well. Finally, compression is not equivalent to
training the final small adapter directly: adapters trained with additional
rank recover more strongly when all are compressed to the same rank-1,
one-bit endpoint. Full controls are reported in
Appendix~\ref{app:recovery-controls}.

\subsection{Damage Is Structured Across Adapter Directions}
\label{subsec:directions}

\begin{figure}[t]
\centering
\includegraphics[width=0.85\linewidth]{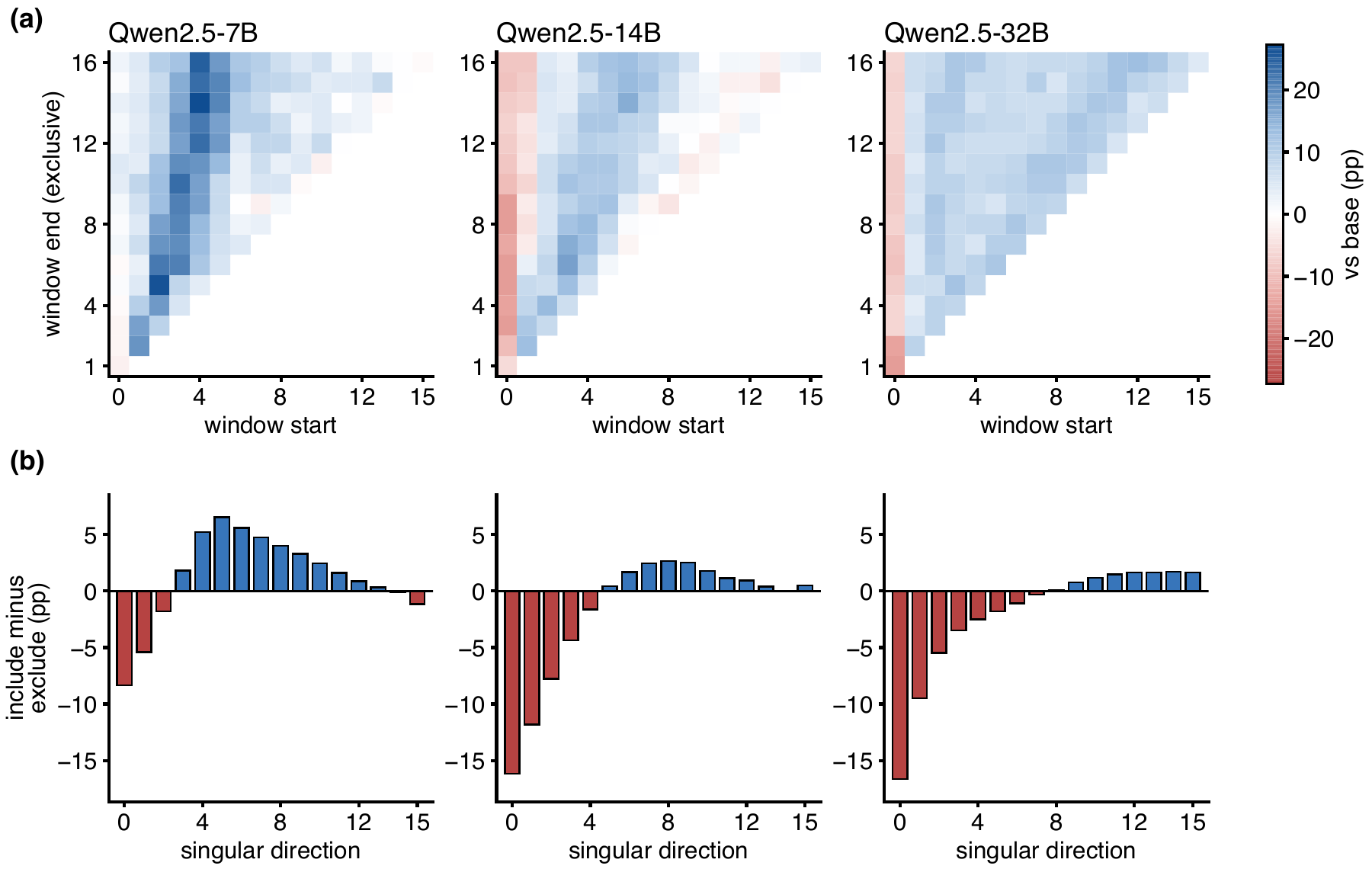}
\caption{
\textbf{Corruption is structured across singular directions.}
(a) Accuracy relative to the frozen model for every contiguous window
$[i,j)$ of singular directions retained at full precision.
(b) Marginal contribution of each direction, computed by comparing windows
that include it with those that do not.
}
\label{fig:r6r7}
\end{figure}

The previous results suggest that the harmful and useful effects of the update
are not uniformly distributed across its rank directions. We test this directly
by decomposing corrupted adapters into singular directions and evaluating every
contiguous window $[i,j)$ of the rank-16 update at full precision. Figure~\ref{fig:r6r7} reveals a consistent qualitative structure across model
scales. Leading directions are strongly detrimental, an intermediate range
contributes positively, and trailing directions have comparatively little
effect. The location and width of these regimes vary across models, but the
behavioral effect is clearly non-uniform across directions. This shows that
the corruption is represented anisotropically within the learned update, and
provides one mechanism by which compression can preferentially attenuate
harmful behavior. The recovery controls above, however, show that directional
selection alone does not account for the full effect.

The effect is not restricted to a single model size or task. Additional
experiments reproduce recovery across Qwen2.5 scales and other corrupted
settings. On 72 off-task probes, the compressed corrupted adapter also remains
closer to the frozen model than a clean fine-tune with comparable in-domain
performance. Scaling, matched-norm, training-rank, cross-model, and off-task
controls are reported in Appendix~\ref{app:recovery-controls}.
Together, these results show that compression need not merely degrade a
fine-tuned update. When supervision induces both useful and harmful effects,
those effects can be distributed very differently across the learned update,
and aggressive bottlenecks can suppress them unequally. In the next section,
we ask whether related properties of a successful update also predict how
much rate is required to preserve its behavior.

\section{What Sets the Behavioral Rate?}
\label{sec:behavior-prediction}

The receiver-relative view of Section~\ref{sec:information_theory} suggests two
complementary questions. First, can compression difficulty be anticipated from
the data and the frozen receiver before training? Second, once an update has
been learned, what property of that update tracks how aggressively it can be
compressed? We address these questions in turn.

\subsection{Behavioral Rate Is Receiver-Relative}
\label{sec:receiver-relative-rate}

\begin{figure}[t]
\centering
\includegraphics[width=0.62\linewidth]{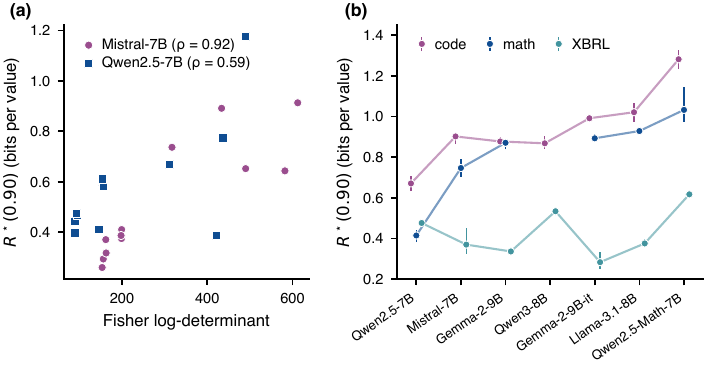}
\caption{
\textbf{Behavioral rate is receiver-relative.}
(a) Correction-spectrum log-determinant against $R^\star(0.90)$ within the
two development receivers.
(b) $R^\star(0.90)$ for the same three corpora across seven receivers; bars
span seeds. The ordering of corpora changes with the receiver.
}
\label{fig:predictor}
\end{figure}

We first test whether behavioral rate can be treated as an intrinsic property
of the fine-tuning data. Figure~\ref{fig:predictor}(b) measures
$R^\star(0.90)$ for the same three corpora across seven frozen receivers.
The ordering changes with the receiver: XBRL tagging is the lowest-rate corpus
for six receivers, but not for Qwen2.5-7B, where math is cheaper. Even closely
related receivers can differ substantially: preserving 90\% of the math gain
requires 1.03 bits per value on Qwen2.5-Math-7B, compared with 0.41 on
Qwen2.5-7B. Behavioral rate therefore depends on the dataset--receiver pair,
rather than on the dataset alone.

Section~\ref{sec:information_theory} motivates a receiver-relative statistic
computed before training from the corrections requested by the data. We
summarize the corresponding correction-gradient spectrum by its
log-determinant (Appendix~\ref{app:correction-statistics}). Within a fixed
receiver, this quantity ranks behavioral rate well: across 24 arms and seven
task families, Spearman correlation with $R^\star(0.90)$ is $\rho=0.92$ on
Mistral-7B and $\rho=0.59$ on Qwen2.5-7B, with a selection-adjusted
permutation $p=0.0003$ (Figure~\ref{fig:predictor}(a)).

This association does not transfer reliably across receivers. After fixing the
statistic, correlations on additional receiver panels are $-0.1$ on
Llama-3.1-8B, $-0.4$ on Qwen3-8B, and $+0.1$ on a fresh Qwen2.5-7B panel.
Corpus-only quantities and quantities obtained by scoring the corpus with the
frozen model likewise fail to consistently order receivers. We therefore view
correction spectra as informative \emph{within-receiver} descriptors of
compression difficulty, rather than as universal predictors of behavioral
rate. Predicting $R^\star$ from the data and receiver alone remains open.

\subsection{Behavioral Rate Tracks Update Shrinkability}
\label{sec:rate-shrinkability}

\begin{figure}[t]
\centering
\includegraphics[width=0.7\linewidth]{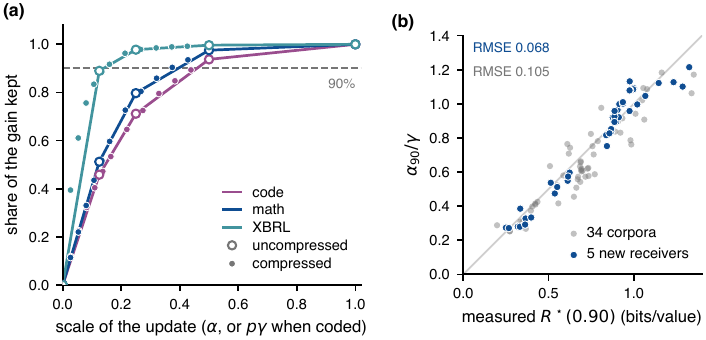}
\caption{
\textbf{Behavioral rate tracks update shrinkability.}
(a) Retained gain for three Llama-3.1-8B adapters when the uncompressed
update is scaled by $\alpha$ (lines), together with compressed operating
points placed at their effective scale $p\gamma$ (dots).
(b) $\alpha_{90}/\gamma$ against measured $R^\star(0.90)$ for 42 checkpoints
on five receivers not used to fix the rule (blue) and 58 archived adapters
spanning 34 corpora (grey). No regression coefficient is fitted.
}
\label{fig:rate-rule}
\end{figure}

The previous analysis asks what can be inferred before the update is trained.
Once the learned update is available, a simpler relationship emerges.
Section~\ref{sec:recovery} showed that aggressive compression often acts
behaviorally like shrinking the update, although it is not exactly equivalent
to scalar rescaling. This suggests that compression difficulty may depend on
how far the learned update can be reduced before its acquired behavior is lost.
\newline
Let $\Delta$ denote the trained update and let
$\widehat{\Delta}_{1\mathrm{bit}}$ denote its one-bit coded version. We define
\begin{equation}
    \gamma
    =
    \frac{
        \langle \widehat{\Delta}_{1\mathrm{bit}},\Delta\rangle
    }{
        \lVert\Delta\rVert_2^2
    },
    \label{eq:gamma}
\end{equation}
which measures how strongly one-bit coding shrinks the update along its
original direction. Under our rank-pair codec, retaining a fraction $p$ of
the directions at one bit behaves approximately like scaling the uncompressed
update by $p\gamma$. Along this compression path, the realized rate is also
approximately $p$ bits per original adapter value.
Figure~\ref{fig:rate-rule}(a) tests this correspondence directly. Across three
Llama-3.1-8B adapters, compressed operating points placed at their effective
scale $p\gamma$ closely follow the behavioral response obtained by explicitly
scaling the uncompressed update by $\alpha$.

Let $\alpha_{90}$ denote the smallest scale at which the uncompressed update
retains 90\% of its original fine-tuning gain. Matching
$p\gamma\approx\alpha_{90}$ then gives the codec-specific approximation
\begin{equation}
    R^\star(0.90)
    \approx
    \frac{\alpha_{90}}{\gamma}.
    \label{eq:rate-rule}
\end{equation}
Estimating $\alpha_{90}$ requires evaluating only five scales of the
uncompressed update on 64 held-out examples, while $\gamma$ is obtained from
the adapter weights; no regression coefficient is fitted.

After fixing the rule on six adapters from two receivers,
Figure~\ref{fig:rate-rule}(b) compares this estimate with measured rates on
42 checkpoints from five other receivers. It achieves an RMSE of
0.068 bits per value ($R^2=0.95$). On a broader archive of 58 adapters
spanning 34 corpora, the RMSE is 0.105 bits per value. Additional baselines,
disjoint-evaluation controls, and trained-update predictors are reported in
Appendix~\ref{app:rate-prediction-controls}.
\newline
These experiments remain retrospective: although the rule was fixed
independently of the reported receiver panels, their behavioral rates had
already been measured when it was formulated. We therefore interpret
Equation~\ref{eq:rate-rule} as a retrospective empirical approximation rather
than a prospectively validated prediction law.
\newline
The same response curve connects the two main empirical findings of the
paper. In Section~\ref{sec:recovery}, moving back along the learned update can
remove harmful behavior and recover performance; here, the distance one can
move before losing useful behavior closely tracks the rate required to
preserve the fine-tune.

\section{Discussion and Limitations}

Our experiments show that behavioral rate is receiver-relative: the same
dataset can require different rates across frozen models, while
parameter-space reconstruction error alone does not predict which compressed
updates preserve behavior. At the same time, the
success of the shrinkability approximation indicates that, once an update has
been learned, much of its compression difficulty can be summarized by the
behavioral response obtained when moving back along that update.
\newline
Several limitations qualify these conclusions. First, $R^\star$ is an
operational, codec-dependent rate rather than an information-theoretic optimum;
different adapter parameterizations or codecs may induce different measured
rates. Second, the correction-spectrum statistic is only a proxy for the full
receiver-relative geometry and does not transfer reliably across receivers.
Third, the shrinkability rule is supported by broad but retrospective
evaluation, and remains to be validated prospectively on rates not observed
when the rule is formulated. Finally, our recovery experiments use deliberately
corrupted supervision to create a controlled mixture of useful and harmful
effects. Whether the same structure governs naturally occurring fine-tuning
errors remains an open question.

\section{Conclusion}

We introduced the behavioral rate of fine-tuning: the smallest serialized
adapter rate that preserves a target fraction of the behavior acquired by
fine-tuning. Across models and tasks, this rate is not determined by parameter
count, weight reconstruction error, or the dataset alone, but depends on the
interaction between the learned update and its frozen receiver.

Compression also exposes structure within the update itself. Under corrupted
supervision, useful and harmful effects can occupy components with sharply
different behavioral consequences, allowing aggressive bottlenecks to recover
performance that the full update destroys. More generally, once an update has
been trained, the rate required to preserve it closely tracks how far that
update can be shrunk before its acquired behavior is lost.

These results separate three notions that are often conflated: how much data
are presented during fine-tuning, how large the resulting parameter update is,
and how many bits of that update are behaviorally necessary for a particular
receiver. Adapter compression therefore serves not only as a storage
mechanism, but as an operational probe of which parts of a learned update are
needed to preserve a specified behavior.

\section*{AI Use Statement}

Generative AI tools were used for language editing and debugging support.
All AI-assisted text were reviewed by the authors. The authors take full
responsibility for the paper's claims, results, citations, and final content.

\section*{Reproducibility Statement}

We provide the experimental details required to reproduce our results in the
appendices. Appendix~\ref{app:xp_details} specifies the models, datasets,
fine-tuning configurations, compression procedures, serialization protocol,
and evaluation setup used to measure behavioral rate. In particular, reported
rates correspond to realized serialized adapter sizes after including the
metadata required for decoding and reloading the adapter. Theoretical
assumptions and derivations for the receiver-relative rate-distortion view are
given in Appendix~\ref{app:rate-distortion}, while additional predictor,
compression, recovery, and generalization controls are reported in the
remaining appendices. We report multiple seeds where indicated in the text and
figure captions, and distinguish explicitly between multi-seed results and
single-checkpoint diagnostic analyses. Code for training, compression, evaluation, and figure generation is available
at \url{https://github.com/joey-david/fineqcomp}.

\subsubsection*{Acknowledgments}
This work was granted access to the HPC resources of IDRIS under the allocation
\mbox{2026-AD011018261} made by GENCI.

\bibliography{biblio}
\bibliographystyle{src/preprint}
\newpage
\appendix
\section{Experimental Details}
\label{app:xp_details}

\subsection{Models, training and evaluation}
\label{app:setup}

\paragraph{Models.} All frozen models are base (pretrained) checkpoints loaded in 4-bit NF4 with double quantization (bitsandbytes), except Mistral-7B-Instruct-v0.3 and Gemma-2-9B-it, which are instruction-tuned and prompted through their chat template. We use Qwen2.5 at 0.5B, 1.5B, 3B, 7B, 14B and 32B parameters, Qwen3-8B-Base, Llama-3.1-8B, Mistral-7B-v0.1, Mistral-7B-Instruct-v0.3 and Gemma-2-9B-it. Most experiments use Qwen2.5-7B. Headline comparisons use three seeds unless stated otherwise; some diagnostic
analyses use a single checkpoint, as indicated below and in the corresponding
figure captions.

\paragraph{Adapters and training.} Unless stated otherwise, adapters are LoRA of rank 16 with scale $\alpha = 32$ ($\alpha/r = 2$) and no dropout, on the attention and MLP projections of every layer (q, k, v, o, gate, up and down). We train with AdamW at learning rate $2\times10^{-4}$, $\beta = (0.9, 0.95)$, no weight decay, 3\% linear warmup, gradient clipping at 1.0, effective batch size 16 (micro-batches of 4) and maximum sequence length 1,024 tokens, with the loss on the whole response. The corruption experiments of Section~\ref{sec:recovery} train for one epoch on 32,000 MetaMathQA rows (2,000 updates), and evaluate the validation loss every 200 updates. The adapters of Figure~\ref{fig:families} and of the 14B and 32B studies keep the checkpoint with the lowest validation loss; the Qwen2.5-7B window-search adapters keep the final one. Seeds are 11, 22 and 33 where three are reported, and 11 otherwise.

\paragraph{Evaluation.} Decoding is greedy. GSM8K-style tasks allow 512 new tokens and are scored by exact match of the final answer; MATH-500 allows 1,024 tokens and is checked with Math-Verify; multiple-choice probes ask for a letter within 32 tokens. GSM8K uses all 1,319 test problems where stated and a fixed subset of 500 otherwise. Intervals are 95\% paired bootstrap intervals over questions (2,000 to 10,000 draws), clustered by template for GSM-Symbolic.

\paragraph{Selecting spectral filters.} For the window searches of Section~\ref{sec:recovery}, the corrupted update is decomposed by SVD into its 16 singular directions. Each candidate keeps a set of directions at full precision or codes them. Candidates are scored on 128 MetaMathQA rows reserved from training (every seed problem of the training slice is excluded), the finalists are re-scored on 128 further reserved rows, and only the selected filter is scored on the test sets.

\subsection{Corruptions}
\label{app:corruptions}
Table~\ref{tab:corruptions} lists the corruptions and how many configurations use each. The main text uses the rationale permutation; the XBRL rows of Table~\ref{tab:scale} use the response mismatch.

\begin{table}[h]
\caption{Corruptions used, with the number of configurations that use each.}
\label{tab:corruptions}
\centering
\small
\begin{tabular}{p{0.3\linewidth}p{0.52\linewidth}r}
\toprule
 & corruption & configs \\
\midrule
rationale permuted & each worked explanation/CoT moved to a different problem; question and answer line kept & 19 \\
response mismatched & the whole response dealt to a different item, for tasks without an answer line & 10 \\
rationale steps shuffled & steps of the explanation reordered; steps, arithmetic and answer kept & 2 \\
arithmetic corrupted & half the computed results moved by up to a third, and the answer follows the wrong chain & 2 \\
labels scrambled / flipped & each multiple-choice letter, or short text label, moved to another of its choices & 2 + 2 \\
\bottomrule
\end{tabular}
\end{table}

\subsection{Codecs and rate accounting}
\label{app:codecs}

Rates are the size of the serialized file, decoding metadata included, divided by the number of scalar values in the uncompressed adapter. The uniform code is a zero-free midrise quantizer fitted per row; the exact-zero control is a midtread quantizer scaled by the row's absolute maximum. LoRAQuant runs in four configurations: 2@0.8, 2@0.9, 3@0.8 and 3@0.9. Below one bit per value we remove paired rank directions before coding. Table~\ref{tab:codecs} lists the rates each code realises.

\begin{table}[h]
\caption{Realised rate and retained GSM8K gain per codec (Mistral-7B, 3 seeds, raw gain 61.5 points).}
\label{tab:codecs}
\centering
\small
\begin{tabular}{lrrr}
\toprule
codec & bits / value & gain kept (\%) & relative RMSE \\
\midrule
midtread, 2 levels (exact zero) & 0.63 & 61.1 & 0.881 \\
adaptive, may drop rows, target 1 & 0.98 & 2.6 & 0.537 \\
uniform zero-free, 1 bit & 1.02 & 85.7 & 0.563 \\
adaptive, keeps rows, target 1.1 & 1.03 & 85.9 & 0.561 \\
LoRAQuant 2@0.8 & 1.59 & 95.4 & -- \\
uniform zero-free, 2 bits & 1.97 & 98.6 & 0.314 \\
LoRAQuant 3@0.9 & 2.69 & 99.1 & -- \\
uniform zero-free, 3 bits & 2.84 & 99.8 & 0.184 \\
uniform zero-free, 4 bits & 3.39 & 100.7 & 0.112 \\
uniform zero-free, 8 bits & 7.28 & 100.1 & 0.008 \\
\bottomrule
\end{tabular}
\end{table}

The compressed adapters of Figure~\ref{fig:families} use the uniform code below one bit: half of the rank directions are removed and the rest are coded at 1 bit, 0.52 bits per value overall and 2.63 to 2.74\,MB depending on the model. The rank-1 codes of Figure~\ref{fig:rank-bits} and Table~\ref{tab:scale} keep the leading singular direction of the update; a rank-1 adapter at 1 bit is a 0.40\,MB file for Qwen2.5-7B.

\subsection{Candidates for the behavioral rate}
\label{app:negative}

\begin{figure}[h]
\centering
\includegraphics[width=\linewidth]{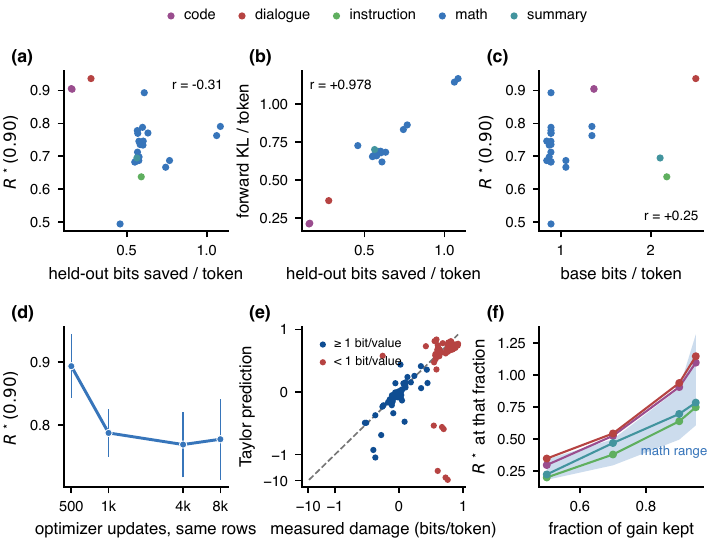}
\caption{Candidates for predictors of $R^\star(0.90)$, one point per arm (seed means; 24 arms, Mistral-7B). (a) Held-out bits saved: $r=-0.31$. (b) Forward KL tracks bits saved at $r=+0.978$, so it is not a second measurement. (c) Base bits per token on the corpus: $r=+0.25$. (d) Sixteen times the optimizer updates on the same rows lowers $R^\star$; bars are seed s.d. (e) Second-order Taylor prediction of the damage against measured damage, symmetric log axes: close above one bit per value, wrong in sign and size below it. (f) $R^\star$ at other retention thresholds: code and dialogue sit at the top of the math range at every threshold.}
\label{fig:a2}
\end{figure}

Figure~\ref{fig:a2} shows the candidates that fail on the 24 Mistral-7B development arms: the size of the behavioral gain, distribution shift, base-model surprisal, optimizer depth, a second-order Taylor model of the damage, and the choice of retention threshold.

\begin{figure}[h]
\centering
\includegraphics[width=0.66\linewidth]{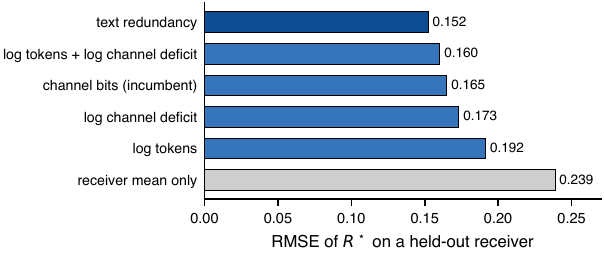}
\caption{Leave-one-receiver-out RMSE of $R^\star(0.90)$ for the rate-law models, four receivers.}
\label{fig:a3}
\end{figure}

The audit re-scored 268 cells from 67 arms on four receivers with 46 candidate measures, after six repairs to the measurement pipeline; nothing was retrained. The repairs lower the held-out-receiver error from 0.239 (receiver mean) to 0.152, but the best predictor is text cross-row redundancy, two zlib calls that never look at the model (Figure~\ref{fig:a3}). A measure of the corpus alone assigns one value to a corpus on every receiver, so it cannot produce the corpus--receiver reorderings of Section~\ref{sec:behavior-prediction}.

Table~\ref{tab:external} lists every test of a correction statistic on a receiver outside the development pair. Only the locked rows were chosen before seeing that receiver’s \(R^\star\); the rest are shown for completeness.

\begin{table}[h]
\caption{Correction statistics on new receivers. $\rho$: Spearman between measure and $R^\star(0.90)$ over natural corpora (seed means). RMSE: absolute prediction from the development fit, against the base-code-length fit. Gate: $\rho\ge0.70$ and lower RMSE than base code length.}
\label{tab:external}
\centering
\small
\setlength{\tabcolsep}{4pt}
\begin{tabular}{llcrrrr}
\toprule
receiver & measure & locked & corpora & $\rho$ & RMSE & base RMSE \\
\midrule
Llama-3.1-8B & Fisher log-volume & yes & 5 & $-0.10$ & 0.357 & 0.327 \\
Qwen3-8B & Fisher effective rank & yes & 4 & $-0.40$ & 0.408 & 0.306 \\
Qwen2.5-7B, new panel & coherent energy / entropy & yes & 5 & 0.10 & 0.296 & 0.241 \\
\midrule
Llama-3.1-8B & coherent energy / entropy & no & 5 & 0.70 & 0.132 & 0.327 \\
Qwen3-8B & coherent energy / entropy & no & 4 & 0.80 & 0.163 & 0.306 \\
Qwen3-8B & coherent energy & no & 4 & 1.00 & 0.206 & 0.306 \\
Llama-3.1-8B & Fisher log-determinant & no & 5 & $-0.50$ & 0.483 & 0.327 \\
Qwen3-8B & Fisher log-determinant & no & 4 & 0.40 & 0.461 & 0.306 \\
Qwen2.5-7B, new panel & Fisher log-determinant & no & 5 & 0.70 & 0.234 & 0.241 \\
\bottomrule
\end{tabular}
\end{table}

\subsection{Known payload}
\label{app:payload}

Sixteen single-token labels are assigned to (family, item) pairs; 128 of 192 mappings are taught, and only the rule that generates labels changes between conditions, so the number of independent 4-bit draws behind them is known. This separates the information the data carries relative to the base model, the weights installed, and the serialized rate by construction.

We additionally estimate the description length of this synthetic source with
a conditional prequential code. After allowing later blocks to fall back to
the frozen model when this gives the shorter code, the estimate correctly
orders the 4-, 64-, 128-, 256-, and 512-bit source conditions and remains
within 1.1-3.5 times the known source description length. The corresponding
behavioral rates do not inherit a simple ordering by source description
length. Thus, even when the amount of source information is controlled and
independently calibrated, it is not sufficient to determine
$R^\star(0.90)$.

\subsection{All generalization probes}
\label{app:breadth}

Table~\ref{tab:a11} lists the 72 probes behind Figure~\ref{fig:breadth}.

{\small\setlength{\tabcolsep}{4pt}
\begin{longtable}{lrrrrrr}
\caption{All 72 generalization probes. Accuracy on all rows; the last two columns are, respectively, the held-out half as a fraction of base, where $\dagger$ marks probes the clean adapter hurts on the first half.}\label{tab:a11}\\
\toprule
probe & tier & base & clean & compr. & clean/base & compr./base \\
\midrule\endfirsthead
\toprule probe & tier & base & clean & compr. & clean/base & compr./base \\ \midrule\endhead
\bottomrule\endfoot
gsm8k & 0 & 0.756 & 0.810 & 0.814 & 1.13 & 1.13 \\
asdiv & 1 & 0.885 & 0.925 & 0.902 & 1.07 & 1.03 \\
gsm symbolic & 1 & 0.762 & 0.830 & 0.790 & 1.10 & 1.09 \\
svamp & 1 & 0.790 & 0.847 & 0.890 & 1.08 & 1.15 \\
aqua rat$^\dagger$ & 2 & 0.402 & 0.059 & 0.394 & 0.02 & 1.00 \\
math500$^\dagger$ & 2 & 0.585 & 0.495 & 0.590 & 0.89 & 1.07 \\
MMLU abstract algebra$^\dagger$ & 2 & 0.470 & 0.320 & 0.510 & 0.70 & 1.15 \\
MMLU college mathematics$^\dagger$ & 2 & 0.510 & 0.170 & 0.470 & 0.34 & 0.90 \\
MMLU elementary mathematics$^\dagger$ & 2 & 0.701 & 0.188 & 0.693 & 0.28 & 0.99 \\
MMLU high school mathematics$^\dagger$ & 2 & 0.556 & 0.011 & 0.419 & 0.01 & 0.69 \\
MMLU high school statistics$^\dagger$ & 2 & 0.690 & 0.449 & 0.671 & 0.63 & 0.97 \\
MMLU anatomy & 3 & 0.681 & 0.696 & 0.674 & 1.00 & 0.98 \\
MMLU astronomy$^\dagger$ & 3 & 0.816 & 0.796 & 0.829 & 1.02 & 1.02 \\
MMLU college biology & 3 & 0.840 & 0.806 & 0.847 & 0.92 & 1.00 \\
MMLU college chemistry$^\dagger$ & 3 & 0.480 & 0.390 & 0.460 & 0.81 & 1.04 \\
MMLU college computer science$^\dagger$ & 3 & 0.650 & 0.550 & 0.640 & 0.84 & 1.03 \\
MMLU college physics$^\dagger$ & 3 & 0.451 & 0.284 & 0.431 & 0.48 & 0.91 \\
MMLU computer security & 3 & 0.810 & 0.840 & 0.810 & 1.05 & 1.00 \\
MMLU conceptual physics$^\dagger$ & 3 & 0.711 & 0.634 & 0.719 & 0.86 & 0.99 \\
MMLU electrical engineering$^\dagger$ & 3 & 0.703 & 0.648 & 0.697 & 0.92 & 1.02 \\
MMLU high school biology & 3 & 0.839 & 0.816 & 0.829 & 0.98 & 0.98 \\
MMLU high school chemistry$^\dagger$ & 3 & 0.665 & 0.463 & 0.645 & 0.71 & 0.99 \\
MMLU high school computer science$^\dagger$ & 3 & 0.800 & 0.720 & 0.780 & 0.87 & 0.92 \\
MMLU high school physics$^\dagger$ & 3 & 0.523 & 0.272 & 0.543 & 0.55 & 1.05 \\
MMLU machine learning$^\dagger$ & 3 & 0.491 & 0.411 & 0.509 & 0.73 & 1.08 \\
arc challenge$^\dagger$ & 3 & 0.880 & 0.825 & 0.887 & 0.94 & 1.02 \\
arc easy & 3 & 0.955 & 0.955 & 0.958 & 0.98 & 0.99 \\
openbookqa & 3 & 0.858 & 0.848 & 0.860 & 0.98 & 1.01 \\
qasc & 3 & 0.823 & 0.807 & 0.830 & 0.98 & 1.00 \\
commonsenseqa & 4 & 0.815 & 0.818 & 0.830 & 1.01 & 1.02 \\
hellaswag$^\dagger$ & 4 & 0.855 & 0.710 & 0.845 & 0.76 & 0.99 \\
winogrande & 4 & 0.650 & 0.637 & 0.632 & 0.97 & 0.96 \\
MMLU business ethics & 4 & 0.750 & 0.720 & 0.740 & 0.91 & 0.94 \\
MMLU clinical knowledge & 4 & 0.770 & 0.751 & 0.766 & 0.94 & 1.00 \\
MMLU college medicine$^\dagger$ & 4 & 0.611 & 0.558 & 0.621 & 0.93 & 1.04 \\
MMLU econometrics & 4 & 0.561 & 0.509 & 0.579 & 0.85 & 1.06 \\
MMLU global facts & 4 & 0.440 & 0.470 & 0.410 & 0.96 & 0.96 \\
MMLU high school geography$^\dagger$ & 4 & 0.889 & 0.859 & 0.884 & 1.00 & 1.01 \\
MMLU high school government and politics & 4 & 0.938 & 0.912 & 0.938 & 0.96 & 1.00 \\
MMLU high school macroeconomics & 4 & 0.772 & 0.751 & 0.764 & 0.99 & 0.99 \\
MMLU high school microeconomics & 4 & 0.857 & 0.853 & 0.840 & 1.00 & 0.99 \\
MMLU high school psychology & 4 & 0.873 & 0.870 & 0.873 & 0.98 & 0.99 \\
MMLU human aging & 4 & 0.731 & 0.722 & 0.758 & 1.01 & 1.02 \\
MMLU human sexuality$^\dagger$ & 4 & 0.832 & 0.771 & 0.809 & 0.91 & 0.96 \\
MMLU management & 4 & 0.922 & 0.903 & 0.903 & 0.96 & 0.96 \\
MMLU marketing & 4 & 0.919 & 0.919 & 0.906 & 1.00 & 0.97 \\
MMLU medical genetics & 4 & 0.820 & 0.780 & 0.790 & 0.88 & 0.90 \\
MMLU miscellaneous & 4 & 0.850 & 0.840 & 0.860 & 0.98 & 1.01 \\
MMLU nutrition & 4 & 0.765 & 0.758 & 0.758 & 1.00 & 0.98 \\
MMLU professional accounting$^\dagger$ & 4 & 0.525 & 0.482 & 0.535 & 0.92 & 1.04 \\
MMLU professional medicine & 4 & 0.746 & 0.739 & 0.746 & 0.99 & 1.03 \\
MMLU professional psychology & 4 & 0.735 & 0.715 & 0.740 & 0.98 & 1.03 \\
MMLU public relations$^\dagger$ & 4 & 0.682 & 0.664 & 0.682 & 1.00 & 1.00 \\
MMLU security studies & 4 & 0.763 & 0.771 & 0.771 & 1.01 & 0.99 \\
MMLU sociology & 4 & 0.856 & 0.861 & 0.861 & 1.00 & 1.02 \\
MMLU us foreign policy$^\dagger$ & 4 & 0.940 & 0.870 & 0.900 & 0.96 & 1.00 \\
MMLU virology$^\dagger$ & 4 & 0.554 & 0.512 & 0.530 & 0.93 & 0.98 \\
race high & 4 & 0.882 & 0.875 & 0.882 & 1.00 & 0.99 \\
humaneval & 5 & 0.421 & 0.415 & 0.427 & 0.86 & 1.00 \\
MMLU formal logic$^\dagger$ & 5 & 0.532 & 0.429 & 0.508 & 0.84 & 1.06 \\
MMLU high school european history & 5 & 0.855 & 0.812 & 0.830 & 0.95 & 0.99 \\
MMLU high school us history & 5 & 0.848 & 0.858 & 0.853 & 1.02 & 1.01 \\
MMLU high school world history & 5 & 0.873 & 0.882 & 0.869 & 1.01 & 1.00 \\
MMLU international law & 5 & 0.810 & 0.793 & 0.818 & 0.96 & 1.00 \\
MMLU jurisprudence & 5 & 0.824 & 0.806 & 0.815 & 1.00 & 1.00 \\
MMLU logical fallacies & 5 & 0.828 & 0.804 & 0.828 & 0.96 & 0.99 \\
MMLU moral disputes & 5 & 0.780 & 0.757 & 0.769 & 0.96 & 0.99 \\
MMLU moral scenarios & 5 & 0.292 & 0.395 & 0.237 & 1.26 & 0.76 \\
MMLU philosophy & 5 & 0.740 & 0.720 & 0.746 & 0.98 & 1.04 \\
MMLU prehistory & 5 & 0.809 & 0.784 & 0.796 & 0.98 & 0.97 \\
MMLU professional law & 5 & 0.490 & 0.475 & 0.460 & 0.97 & 0.94 \\
MMLU world religions & 5 & 0.860 & 0.877 & 0.860 & 1.01 & 0.99 \\
\end{longtable}

}

\section{Details on the Receiver-Relative Rate-Distortion View}
\label{app:rate-distortion}

This appendix provides additional details for the local arguments used in
Section~\ref{sec:information_theory}.
The purpose of these derivations is not to establish an exact rate-distortion
theory of LoRA, but to make explicit the assumptions under which the geometric
quantities introduced in the main text arise.

\subsection{Local Behavioral Distortion}
\label{app:local-distortion}

Let $A$ denote a trained adapter and let $\hat A=A+e$ be a perturbed version,
where $e$ is the error induced by compression.
For a differentiable evaluation loss $\mathcal{L}_U$, a Taylor expansion around
$A$ gives
\begin{equation}
\begin{split}
    \mathcal{L}_U(M\oplus(A+e))
    -
    \mathcal{L}_U(M\oplus A)
    &=
    g_U^\top e
    +
    \frac{1}{2}e^\top H_Ue
    +
    o(\|e\|^2),
\end{split}
\label{eq:app-taylor}
\end{equation}
where
\[
    g_U
    =
    \nabla_A \mathcal{L}_U(M\oplus A),
    \qquad
    H_U
    =
    \nabla_A^2 \mathcal{L}_U(M\oplus A).
\]

The quadratic approximation used in the main text therefore requires the
first-order contribution to be negligible for the perturbations considered.
This is exact at a stationary point of the evaluation objective and can also
be a useful local approximation when $|g_U^\top e|$ is small compared with the
quadratic term.

The exact Hessian need not be positive semidefinite.
When interpreting $H_U$ as a local distortion metric, one may instead use a
positive-semidefinite curvature approximation, such as a generalized
Gauss--Newton matrix.
The relevant point for our argument is not the particular approximation, but
that behavioral distortion generally defines a non-isotropic metric over
parameter perturbations.

\subsection{When Does Weight Error Recover Functional Distortion?}
\label{app:isotropy}

Squared weight reconstruction error corresponds to the Euclidean quadratic
form
\[
    D_{\mathrm{weight}}(e)=e^\top e,
\]
whereas the local functional criterion is
\[
    D_{\mathrm{func}}(e)=e^\top H_Ue.
\]

Let $\mathcal{S}$ denote the subspace containing the compression errors under
consideration.
If
\begin{equation}
    H_U|_{\mathcal{S}}=\lambda I,
    \qquad \lambda>0,
    \label{eq:isotropic-condition}
\end{equation}
then
\[
    D_{\mathrm{func}}(e)
    =
    \lambda D_{\mathrm{weight}}(e)
\]
for every $e\in\mathcal{S}$, and the two criteria induce identical rankings of
compression errors.

Conversely, if all perturbations with equal Euclidean norm have equal
functional cost, the restriction of $H_U$ to $\mathcal{S}$ must be proportional
to the identity.
To see this, diagonalize the symmetric quadratic form on $\mathcal{S}$.
If two eigenvalues differ, perturbations of equal norm along their associated
eigenvectors have different functional costs.

Thus, ordinary reconstruction error is sufficient only under an isotropic
functional geometry on the errors produced by the codec.
The RMSE counterexample and layer-location controls in the main text provide
empirical evidence against this approximation in our setting.

\subsection{Correction Statistics and the Learned Update}
\label{app:correction-statistics}

At the frozen parameters $\theta_0$, let
\[
    g_i=\nabla_\theta\ell_i(\theta_0),
    \qquad
    \mu=\mathbb{E}[g_i],
    \qquad
    J=\mathbb{E}[g_ig_i^\top].
\]
The centered correction covariance is
\begin{equation}
    C_g
    =
    \mathbb{E}\big[
        (g_i-\mu)(g_i-\mu)^\top
    \big]
    =
    J-\mu\mu^\top.
    \label{eq:app-centered-gradient}
\end{equation}

A quadratic approximation to the average training objective has the form
\[
    \mathcal{L}_T(\theta_0+\Delta\theta)
    \approx
    \mathcal{L}_T(\theta_0)
    +
    \mu^\top\Delta\theta
    +
    \frac{1}{2}
    \Delta\theta^\top H_T\Delta\theta.
\]
When $H_T$ is invertible on the relevant subspace, its minimizer is
\begin{equation}
    \Delta\theta^\star
    =
    -H_T^{-1}\mu.
    \label{eq:app-newton-update}
\end{equation}

Throughout, $H_T^{-1}$ denotes the inverse restricted to the relevant
subspace. When $H_T$ is singular, the same local expressions can instead be
read using a pseudoinverse or a suitably damped inverse.

Equation~\ref{eq:app-newton-update} makes clear that $J$ does not determine the
deterministic local update.
For example, in one dimension, a dataset for which every example has gradient
$+1$ and a dataset containing equal numbers of gradients $+1$ and $-1$ both
satisfy
\[
    J=1,
\]
but their means are respectively $\mu=1$ and $\mu=0$.
Their local optimal updates are therefore different despite having identical
second moments.

We consequently interpret $J$ as a descriptor of the geometry of the
corrections requested by the data, rather than as an estimate of the adapter
update itself.
Its potential relevance to compression comes from the directional diversity
that it captures in addition to the mean correction.

\subsection{From Correction Statistics to a Receiver-Relative Spectrum}
\label{app:relative-spectrum}

To motivate a link between correction statistics and compressibility, consider
the following local linear model.
Let $G$ be a random correction gradient and suppose that it induces a local
response
\begin{equation}
    \Delta\theta=-H_T^{-1}G.
    \label{eq:app-linear-response}
\end{equation}
This should be understood as an idealized response model, not as a description
of an SGD trajectory.

Under Equation~\ref{eq:app-linear-response}, the mean update is
\[
    \mathbb{E}[\Delta\theta]
    =
    -H_T^{-1}\mu,
\]
its covariance is
\begin{equation}
    \mathrm{Cov}(\Delta\theta)
    =
    H_T^{-1}C_gH_T^{-1},
    \label{eq:app-update-covariance}
\end{equation}
and its second moment is
\begin{equation}
    \mathbb{E}[
        \Delta\theta\Delta\theta^\top
    ]
    =
    H_T^{-1}JH_T^{-1}.
    \label{eq:app-update-second-moment}
\end{equation}

Behavioral distortion introduces a second geometry through $H_U$.
Applying it to the centered update covariance in
Equation~\ref{eq:app-update-covariance} gives
\begin{equation}
    C_{\mathrm{rel}}
    =
    H_U^{1/2}
    H_T^{-1}
    C_g
    H_T^{-1}
    H_U^{1/2}.
    \label{eq:app-relative-covariance}
\end{equation}
This is the quantity that corresponds most directly to the Gaussian
rate-distortion argument below: it is the covariance of the local update
distribution expressed in behaviorally weighted coordinates.

For prediction, however, we are also interested in the systematic correction
encoded by $\mu$. Since
\[
    J=C_g+\mu\mu^\top,
\]
retaining the uncentered second moment motivates the related geometry
\begin{equation}
    \widetilde C_{\mathrm{rel}}
    =
    H_U^{1/2}
    H_T^{-1}
    J
    H_T^{-1}
    H_U^{1/2}.
    \label{eq:app-relative-second-moment}
\end{equation}
Unlike $C_{\mathrm{rel}}$, $\widetilde C_{\mathrm{rel}}$ is not itself a
Gaussian source covariance. It is a second-moment extension that retains both
the mean correction and its directional variability, and is introduced as a
candidate geometry for predicting behavioral rate.

When training and evaluation share the same local geometry,
$H_T=H_U=H$, this reduces to
\begin{equation}
    \widetilde C_{\mathrm{rel}}
    =
    H^{-1/2}JH^{-1/2}.
    \label{eq:app-whitened-spectrum}
\end{equation}

This whitening has a simple interpretation.
A correction direction is discounted when the training geometry makes it easy
to realize, and amplified when it corresponds to a behaviorally sensitive
direction of the receiver.

\subsection{Gaussian Rate-Distortion Motivation}
\label{app:gaussian-rd}

The spectral summaries used in the main text are motivated by the classical
rate-distortion function of a Gaussian source under quadratic distortion
\citep[Ch.~10]{cover2006elements}.

Let
\[
    X\sim\mathcal{N}(0,\Sigma)
\]
and consider a quadratic distortion
\[
    d(x,\hat x)
    =
    (x-\hat x)^\top H(x-\hat x),
    \qquad H\succeq0.
\]
After the change of coordinates
\[
    Z=H^{1/2}X,
\]
the problem becomes ordinary squared-error coding for a Gaussian variable with
covariance
\[
    C=H^{1/2}\Sigma H^{1/2}.
\]
Let $\lambda_1,\ldots,\lambda_d$ denote the eigenvalues of $C$.
The Gaussian rate-distortion function is then given by reverse water filling:
\begin{equation}
    R(D)
    =
    \frac{1}{2}
    \sum_j
    \left[
        \log_2\frac{\lambda_j}{\nu}
    \right]_+,
    \label{eq:app-waterfilling}
\end{equation}
where $\nu$ is chosen so that
\[
    D
    =
    \sum_j\min(\lambda_j,\nu).
\]

Equation~\ref{eq:app-waterfilling} highlights two properties relevant to our
setting.
Rate increases when the distortion-weighted variance becomes larger, but also
when it is spread across more active directions.
This motivates considering both scale-sensitive and
dimensionality-sensitive summaries of correction spectra.

A convenient smooth summary is
\begin{equation}
    I_{\mathrm{rel}}(\alpha)
    =
    \frac{1}{2}
    \log\det
    \left(
        I+\alpha C_{\mathrm{rel}}
    \right)
    =
    \frac{1}{2}
    \sum_j
    \log(1+\alpha\lambda_j).
    \label{eq:app-soft-logdet}
\end{equation}
Unlike Equation~\ref{eq:app-waterfilling}, this expression does not implement
an explicit distortion threshold and should not be interpreted as the
rate-distortion function of our adapters.
We use it only as a smooth statistic that increases with both the magnitude
and dimensionality of the relevant spectrum.

Our empirical correction spectrum is a substantially coarser approximation.
It is constructed from projected output-head correction gradients rather than
from gradients in the full LoRA parameter space, and we do not explicitly
estimate $H_T$ or $H_U$ in the experiments reported here.
Consequently,
\[
    \log\det(I+\alpha J)
\]
should be interpreted as a tractable proxy for the second-moment geometry
$\widetilde C_{\mathrm{rel}}$, which is itself an extension of the covariance
geometry directly motivated by Gaussian rate-distortion theory.
It is therefore neither a direct estimate of
Equation~\ref{eq:app-relative-covariance} nor an information-theoretic rate.

\section{Additional Analyses of Behavioral Rate}
\label{app:behavioral-rate-controls}

This appendix collects additional controls supporting the behavioral-rate
analysis in Sections~\ref{sec:diagnostics} and
\ref{sec:behavior-prediction}. We first summarize alternative quantities that
fail to determine $R^\star(0.90)$, then examine the effect of where a fixed
compression budget is spent, and finally provide additional receiver-level
results.

\subsection{Behavioral Rate and Distinct Training Content}
\label{app:content-diversity}

Figure~\ref{fig:compression-diagnostics} isolates two forms of training-content diversity.
First (left), we vary the number of distinct MetaMathQA rows while holding the
optimization budget fixed , so that larger datasets are seen fewer times.
Across 2k--32k distinct rows, $R^\star(0.90)$ increases with the amount of
non-repeated training content rather than with the number of optimization
updates.

Second (center), we hold the dataset size fixed at 8,000 rows while varying how many
distinct source problems generated those rows. Increasing source-problem
diversity again increases the measured behavioral rate. This indicates that
the effect is not explained solely by the number of stored rows or sample
presentations.

These controls should be interpreted within a fixed corpus and receiver.
Distinct-example count does not provide a universal predictor of
$R^\star$ across unrelated corpora; Table~\ref{tab:not-rstar} reports this
failure together with other alternative predictors.

\subsection{Alternative Predictors of Behavioral Rate}
\label{app:alternative-rate-predictors}

Table~\ref{tab:not-rstar} summarizes several intuitive candidates that do not
by themselves determine $R^\star(0.90)$. Some correlate with rate within
restricted experimental families, but none consistently orders compression
difficulty across datasets and receivers.

\begin{table}[t]
\caption{
\textbf{Alternative quantities do not determine $R^\star(0.90)$.}
Results cover 81 adapters over five corpora (MetaMathQA, Magicoder, XSum,
hh-rlhf, Alpaca) on Mistral-7B and Qwen2.5-7B, together with the broader
rate-law audit described in Appendix~\ref{app:xp_details}.
}
\label{tab:not-rstar}
\centering
\small
\begin{tabular}{p{0.34\linewidth}p{0.58\linewidth}}
\toprule
candidate & result \\
\midrule
distinct rows
& ranks four conditions correctly within a corpus, but reverses across corpora \\

behavioral change (bits saved)
& slope 1.90 within a 0.16-wide window; 0.12 on answer rewrites and approximately flat under optimizer-budget changes \\

optimizer depth
& increasing the number of updates by $16\times$ on the same rows changes bits saved by 0.04 while $R^\star$ decreases \\

base surprisal on the corpus
& $r=+0.25$ with $R^\star$ \\

weight reconstruction error
& approximately constant to within 1\% across all five corpora at a fixed rate \\

local curvature
($g^\top\delta$, $\tfrac12\delta^\top H\delta$)
& the second-order term adds no ranking power, and the local expansion breaks down below one bit per value \\
\bottomrule
\end{tabular}
\end{table}

\subsection{Equal Bit Budgets Are Not Functionally Interchangeable}
\label{app:layer-location}

Section~\ref{sec:diagnostics} shows that the magnitude of parameter-space
reconstruction error is not sufficient to predict behavioral distortion.
Here we additionally test whether the location of that error matters.

\begin{figure}[t]
\centering
\includegraphics[width=0.8\linewidth]{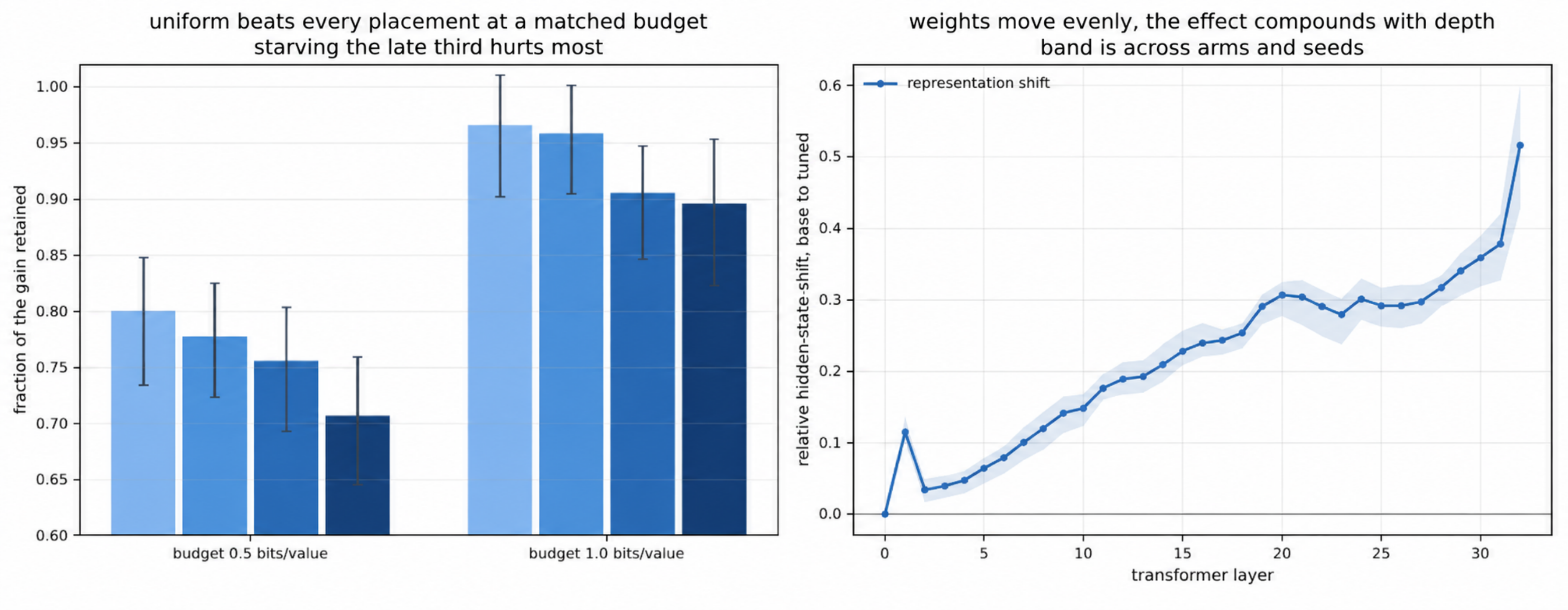}
\caption{
\textbf{Equal bit budgets are not functionally interchangeable.}
Left: retained gain under uniform or depth-selective compression at matched
budgets. Right: hidden-state shift grows with depth despite approximately
flat update magnitude.
}
\label{fig:layer-location}
\end{figure}

We compare uniform precision with matched-rate variants in which precision is
reduced preferentially in the early, middle, or late third of the transformer.
Figure~\ref{fig:layer-location} shows that uniform allocation preserves more
fine-tuning gain at both 0.5 and 1.0 bits per value, while reducing precision
in the final third is consistently the most damaging.

This difference is not explained by larger parameter updates in later layers.
The RMS magnitude of the learned update is approximately flat across depth,
whereas the relative hidden-state shift induced by fine-tuning grows toward
the final layers. Thus, comparable parameter-space perturbations can have
different behavioral effects depending on where they act in the frozen model.

\subsection{Alternative Trained-Update Predictors}
\label{app:rate-prediction-controls}

The main text distinguishes between quantities available before training and
properties of the learned update itself. Here we report additional
trained-update predictors and controls omitted from the main discussion.

A simple candidate is the fraction of update energy carried by the leading
singular direction. We fixed this statistic before evaluating a confirmation
panel of 64 adapters spanning four receivers and 34 corpora. On held-out
receivers, it orders receiver pairs correctly 73\% of the time, compared with
48\% for the best correction-spectrum statistic, and reduces prediction error
from 0.167 bits per value for a corpus-only baseline to 0.129.

We also tested a more behaviorally informed directional statistic that measures
how much training-data gain is carried by individual singular directions.
Despite using both the trained adapter and the receiver's behavior on the
training data, this statistic does not transfer reliably across receivers,
achieving 17\% receiver-pair sign accuracy in the same evaluation.

These results motivated the shrinkability rule used in
Section~\ref{sec:rate-shrinkability}. For completeness, we report additional
validation of
\[
    R^\star(0.90) \approx \alpha_{90}/\gamma.
\]
On rates remeasured using evaluation rows disjoint from those used to estimate
$\alpha_{90}$, the rule obtains an RMSE of 0.048 bits per value. On the five
new-receiver panel and the broader 34-corpus archive, the corresponding RMSEs
are 0.068 and 0.105, compared with 0.093, 0.076, and 0.177 for the strongest
alternative baselines in the respective evaluations.

Because the behavioral rates in these panels had already been measured when
the shrinkability rule was formulated, these experiments remain retrospective.
We therefore use them as evidence that update shrinkability is a strong
empirical descriptor of behavioral rate, rather than as a prospectively
validated prediction law.

\section{Additional Controls for Compression-Based Recovery}
\label{app:recovery-controls}

This appendix provides additional controls for
Section~\ref{sec:recovery}. We test alternative compression operations,
separate coarse coding from scalar shrinkage, vary the rank available during
training, replicate the directional analysis across model families, and
evaluate recovery across model scales, tasks, and out-of-distribution probes.

\subsection{Recovery under Alternative Compression Operations}
\label{app:recovery-operations}

The main recovery experiments combine aggressive rank reduction with coarse
quantization. We first test whether recovery is specific to this particular
compression path. Table~\ref{tab:ladder} compares several alternative
interventions on the same corrupted Qwen2.5-7B adapters.

The rank-1, one-bit representation gives the strongest recovery in this
comparison, but neither rank reduction nor quantization alone is sufficient
to explain the phenomenon. In particular, a random rank-1 projection at full
precision also recovers substantial performance, whereas simply scaling the
full update to the norm of the rank-1 update does not.

\begin{table}[t]
\caption{
\textbf{Recovery under alternative compression operations.}
Accuracy (\%) on GSM8K (1,319 examples) and GSM-Symbolic
(500 examples, intervals clustered by template), averaged over three seeds.
Frozen-base accuracy is 77.56 / 71.40.
}
\label{tab:ladder}
\centering
\small
\begin{tabular}{lrr}
\toprule
condition & GSM8K & GSM-Symbolic \\
\midrule
rank 1 @ 1 bit
& \textbf{83.24} & \textbf{77.60} \\

rank 1 @ 16 bits
& 42.81 & 33.80 \\

full rank @ 1 bit
& 65.05 & 55.00 \\

full rank @ 16 bits (damaged)
& 23.81 & 15.53 \\

random rank-1 projection @ 16 bits
& 80.09 & 72.13 \\

full update scaled to the rank-1 norm
& 23.93 & 13.80 \\

residual rank 15 @ 16 bits
& 30.86 & 25.20 \\
\bottomrule
\end{tabular}
\end{table}

Uniformly coding the complete adapter, without removing rank directions,
already repairs part of the damage while having a comparatively small effect
on an aligned adapter. For example, at one bit per value the corrupted
Qwen2.5-7B adapter rises from 23.7\% to 65.8\% GSM8K accuracy, whereas the
aligned adapter changes from 83.2\% to 82.4\%.

\begin{table}[t]
\caption{
\textbf{Uniform low-bit coding can itself recover corrupted adapters.}
GSM8K accuracy (\%) of aligned and permuted adapters under three uniform
codes, averaged over three seeds. Realized rates are 1.97, 1.50, and
1.02 bits per value.
}
\label{tab:a6}
\centering
\small
\begin{tabular}{lrrrrrrrr}
\toprule
& \multicolumn{4}{c}{aligned}
& \multicolumn{4}{c}{permuted} \\
\cmidrule(lr){2-5}\cmidrule(lr){6-9}
model
& raw & 2 bits & 1.5 bits & 1 bit
& raw & 2 bits & 1.5 bits & 1 bit \\
\midrule
Mistral-7B
& 68.9 & 67.3 & 64.2 & 59.3
& 15.7 & 15.3 & 14.3 & 24.2 \\

Qwen2.5-7B
& 83.2 & 83.4 & 82.9 & 82.4
& 23.7 & 23.2 & 27.4 & 65.8 \\

Llama-3.1-8B
& 73.2 & 72.2 & 70.0 & 65.7
& 19.8 & 19.4 & 17.1 & 34.7 \\
\bottomrule
\end{tabular}
\end{table}

\subsection{Shrinkage and Training-Rank Controls}
\label{app:recovery-shrinkage}

A natural explanation for recovery is that coarse quantization simply reduces
the magnitude of the corrupted update. We therefore compare coded rank-1
updates with the corresponding full-precision rank-1 update rescaled by a
scalar $\alpha$.

\begin{figure}[t]
\centering
\includegraphics[width=0.8\linewidth]{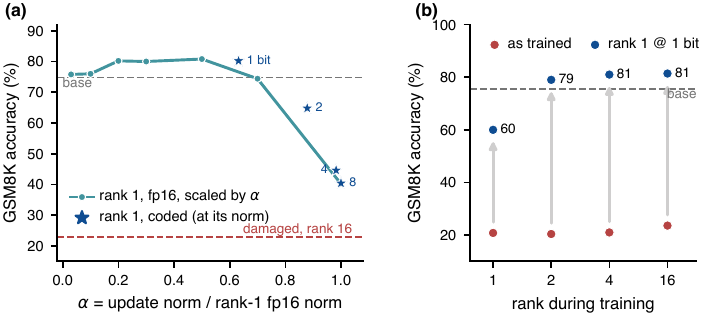}
\caption{
\textbf{Shrinkage and training-rank controls.}
(a) GSM8K accuracy of the full-precision rank-1 update as it is rescaled by
$\alpha$, compared with coded rank-1 updates at their realized norms.
(b) GSM8K accuracy after corrupted adapters trained at different ranks are all
compressed to the same rank-1, one-bit endpoint.
}
\label{fig:shrinkage}
\end{figure}

Figure~\ref{fig:shrinkage}a shows that shrinkage explains a substantial
fraction of the recovery: an appropriately rescaled full-precision update can
approach the performance of the one-bit code. However, matched-norm controls
do not eliminate all differences between the coded and rescaled updates,
indicating that coarse coding is not exactly equivalent to scalar shrinkage.

\paragraph{Effect of training rank.}
The final recovered adapter is rank 1 and one bit, raising the possibility
that the same result could be obtained by training a small adapter from the
start. We therefore train corrupted adapters at ranks 1, 2, 4, and 16, and
compress every checkpoint to the same rank-1, one-bit endpoint.
Figure~\ref{fig:shrinkage}b shows that recovery is substantially stronger when
additional rank was available during training: final GSM8K accuracy rises
from approximately 60\% when trained at rank 1 to approximately 78--81\% when
trained at ranks 2--16.

Compression is therefore not equivalent to directly training the final
low-rank representation. Higher training rank appears to make useful behavior
more recoverable under the subsequent bottleneck, although these experiments
do not establish the mechanism by which this separation emerges.

\subsection{Directional Structure across Model Families}
\label{app:directional-replication}

\begin{figure}[t]
\centering
\includegraphics[width=\linewidth]{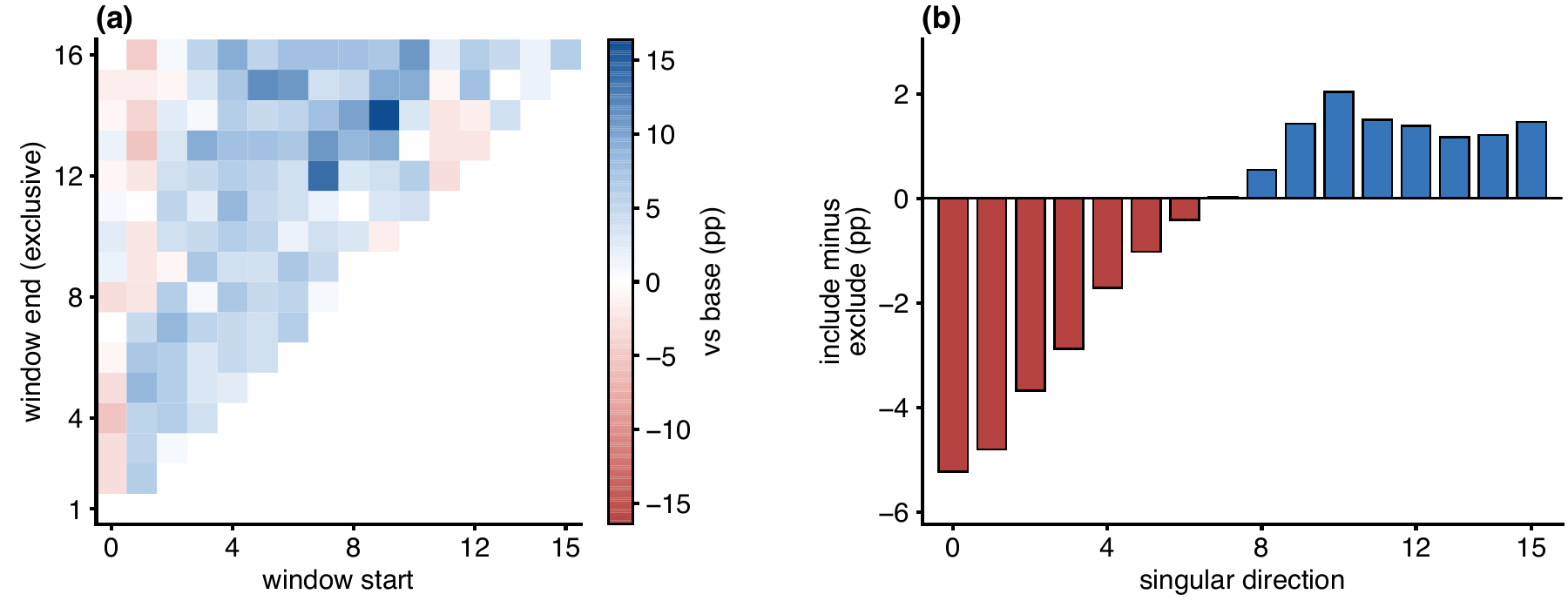}
\caption{
\textbf{Directional structure on Mistral-7B-Instruct.}
(a) Accuracy relative to the frozen model for every contiguous window
$[i,j)$ of singular directions retained at full precision.
(b) For each direction, mean accuracy of windows containing it minus that of
windows excluding it. Results use 128 reserved MetaMathQA examples.
}
\label{fig:passband-mistral}
\end{figure}

We repeat the singular-direction analysis of
Section~\ref{subsec:directions} on Mistral-7B-Instruct in
Figure~\ref{fig:passband-mistral}. The same qualitative structure appears as
for Qwen2.5: windows containing the leading directions are generally harmful,
while windows beginning later in the spectrum recover task performance.

Every window containing the leading direction remains at or below the frozen
baseline (24.0\% on average versus 25.0\% for the base), whereas windows
starting between directions 2 and 9 average 28.7--31.8\%. Consistently, the
per-direction marginal contribution changes from negative for early
directions to positive for later ones.

\subsection{Recovery across Model Scales and Tasks}
\label{app:recovery-scale}

We next test whether recovery is specific to Qwen2.5-7B and GSM8K.
Table~\ref{tab:scale} reports corresponding experiments across Qwen2.5 model
scales and on a second structured task.

For models up to 7B, corrupted MetaMathQA rationales consistently damage
GSM8K accuracy and the compressed adapter recovers a substantial fraction of
the loss. For the 14B and 32B models, we additionally report XBRL tagging
under corrupted response assignments, where the frozen models leave more
headroom. Recovery remains visible in this setting.

\begin{table}[t]
\caption{
\textbf{Recovery across model scales and tasks.}
For models up to 7B, entries are GSM8K accuracy (\%) after training on
permuted MetaMathQA rationales. For 14B and 32B, entries are XBRL exact match
after corrupted response assignments.
}
\label{tab:scale}
\centering
\small
\begin{tabular}{lrrrrr}
\toprule
size & base & damaged & clean & rank 1 @ 1 bit & overshoot \\
\midrule
0.5B
& 23.0 & 9.4 & 45.2 & 25.0 & +2.0 \\

1.5B
& 51.6 & 13.6 & 67.6 & 59.0 & +7.4 \\

3B
& 60.4 & 17.4 & 74.0 & 71.0 & \textbf{+10.6} \\

7B
& 75.6 & 22.8 & 81.0 & 80.8 & +5.2 \\

\midrule

14B, XBRL tagging
& 24.6 & 53.0 & -- & 73.8 & +49.2 \\

32B, XBRL tagging
& 32.4 & 51.8 & -- & 74.4 & +42.0 \\
\bottomrule
\end{tabular}
\end{table}

The XBRL corruption differs from the permuted-rationale intervention:
mismatched responses can still teach the model task-specific output
vocabulary, so the corrupted adapter may already outperform the frozen model.
The relevant comparison is therefore whether compression further separates
task-relevant structure from the corruption-induced behavior.

\subsection{Off-Task Behavior after Recovery}
\label{app:recovery-breadth}

Finally, we ask whether recovering in-domain performance comes at the cost of
larger behavioral changes elsewhere. We evaluate the frozen model, a clean
fine-tune, and the compressed corrupted adapter on 72 probes of varying
proximity to GSM8K.

\begin{figure}[t]
\centering
\includegraphics[width=0.75\linewidth]{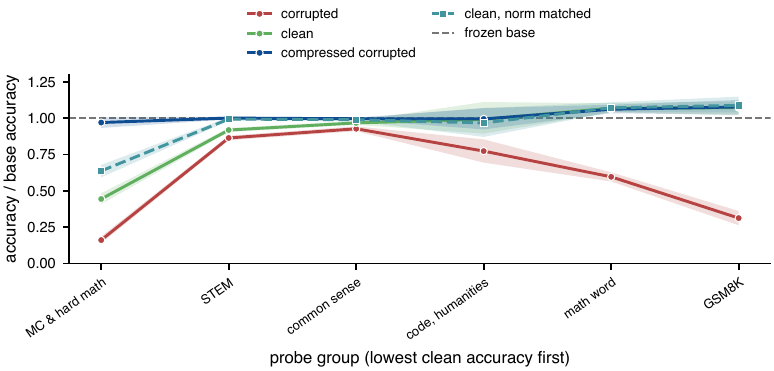}
\caption{
\textbf{Off-task behavior after compression-based recovery.}
Accuracy on 72 probes normalized by the frozen Qwen2.5-7B model and grouped
by task proximity to GSM8K. The compressed corrupted adapter remains close to
the frozen model away from its training task, unlike the clean fine-tune even
under a norm-matched control.
}
\label{fig:breadth}
\end{figure}

Figure~\ref{fig:breadth} shows that the compressed corrupted adapter remains
close to the frozen model on distant task groups despite recovering strong
GSM8K performance. The contrast is particularly large on harder mathematical
tasks: the clean adapter retains 44\% of the frozen model's accuracy, whereas
the compressed corrupted adapter retains 97\%. A norm-matched clean control
indicates that this difference is not explained solely by the magnitude of
the learned update.

\section{Additional Related Work}
\label{app:related-work}

\paragraph{Low-rank parameterizations and capacity.}
A large body of work has refined the original LoRA parameterization by
changing how adaptation capacity is allocated or represented.
AdaLoRA adaptively allocates a parameter budget across weight matrices and
prunes less important singular components of the learned update
\citep{zhang2023adalora}. VeRA substantially reduces trainable storage by
sharing frozen low-rank matrices across layers and learning only small scaling
vectors \citep{kopiczko2024vera}, while LoRA-XS uses frozen SVD-derived bases
and trains only a small $r\times r$ matrix between them
\citep{balazy2024lora}. PiSSA instead initializes the adaptive subspace from
the principal singular directions of the pretrained weights
\citep{meng2024pissa}. In the opposite direction, MoRA argues that restricting
the update itself to low rank can limit learning and memorization, and obtains
higher-rank updates under a comparable parameter budget
\citep{jiang2024mora}. These methods demonstrate that trainable parameter
count, update rank, and the subspace in which adaptation occurs are distinct
design choices. Our behavioral rate measures a different quantity: how many
serialized bits of a trained update must survive in order to preserve its
behavior when composed with a particular frozen receiver.

\paragraph{Optimization and parameterization of LoRA updates.}
Other work shows that even when the represented update class is similar, its
factorization and optimization can substantially affect learning. LoRA+
assigns different learning rates to the two LoRA factors to improve feature
learning \citep{hayou2024loraplus}, while LoRA-FA freezes one factor to reduce
activation-memory cost \citep{zhang2023lora}. DoRA decomposes pretrained
weights into magnitude and direction and applies low-rank adaptation to the
directional component \citep{liu2024dora}. More recently, Balanced LoRA
observes that many pairs of LoRA factors represent the same weight update but
can have very different conditioning, and projects optimization onto a
balanced factorization \citep{castin2026balanced}. This distinction between
the parameterization used to learn an update and the functional effect of the
resulting update is complementary to our setting: our compression analysis is
performed after training and evaluates the decoded update through the frozen
model rather than through its factorization alone.

\paragraph{Selecting and composing adapter components.}
Several methods exploit non-uniform functional importance within or across
adapters. LoRA-drop ranks layers by their LoRA output contribution, retains separate modules for important layers, and shares LoRA parameters across the remaining layers \citep{zhou2025lora}, while LoraHub composes
independently trained LoRA modules to transfer across tasks
\citep{huang2024lorahub}. These approaches further illustrate that adapter
components are not interchangeable. Our experiments ask a complementary
question: after an adapter has been learned, which parts of its representation
must remain after compression to preserve the acquired behavior?

\paragraph{Reasoning structure and chain-of-thought compression.}
Several of our corruption experiments use chain-of-thought supervision
\citep{wei2022cot}. \citet{qian2025reasoningdynamics} study information
dynamics along reasoning trajectories, while \citet{dutta2024stepbystep}
analyze mechanisms supporting step-by-step reasoning. Related mechanistic
work studies symbolic computation and structured state tracking
\citep{yang2025symbolicmechanisms,fagnou2024chainandcausal}.
\citet{li2026slowthinkfast} directly compress chain-of-thought traces using
step-level information measures. Our experiments instead use corrupted
reasoning supervision as a controlled way to induce useful and harmful
effects within the same learned update.

\end{document}